\pdfoutput=1

\documentclass[11pt]{article}
\usepackage[final]{acl} 
\usepackage{times}
\usepackage{latexsym}
\usepackage{amsmath}
\usepackage{amssymb}
\usepackage{graphicx}
\usepackage{booktabs} 
\usepackage{microtype} 
\usepackage{url}
\usepackage[T1]{fontenc}
\usepackage[utf8]{inputenc}
\usepackage{microtype}
\usepackage{inconsolata}
\usepackage{graphicx}
\usepackage{graphicx}
\usepackage{subcaption}
\usepackage{changepage} 
\usepackage{changepage}      
\usepackage{caption}         
\usepackage{subcaption}
\usepackage{cleveref}
\newcommand{\indicconformer}{\texttt{indicconformer}}
\newcommand{\indicSUPERB}{\texttt{IndicSUPERB}}
\newcommand{\ewc}{EWC}
\newcommand{\mas}{MAS}
\newcommand{\lwf}{LwF}
\newcommand{\rnnt}{RNNT}
\newcommand{\ctc}{CTC}
\newcommand{\asr}{ASR}
\newcommand{\cl}{CL}

\usepackage{amsmath} 
\usepackage{enumitem} 

\title{A Study on Regularization-Based Continual Learning Methods for Indic ASR}

\author{
  Gokul Adethya T \textnormal{ and } Dr. S. Jaya Nirmala \\
  Department of Computer Science and Engineering \\
  National Institute of Technology, Tiruchirappalli
}

\date{}

\begin{document}
\maketitle

\begin{abstract}
India's linguistic diversity challenges inclusive Automatic Speech Recognition (\asr) system development. Traditional multilingual models, requiring simultaneous access to all language data, are impractical due to sequential data arrival and privacy constraints. Continual Learning (\cl) enables models to learn new languages sequentially without catastrophically forgetting prior knowledge. This paper investigates \cl\ for \asr\ on Indian languages using the subset of the \indicSUPERB\ benchmark. We employ a Conformer-based hybrid \rnnt-\ctc\ model, initially pretrained on Hindi, which is subsequently trained incrementally on eight additional Indian languages, for a sequence of nine languages in total. We evaluate three prominent regularization and distillation-based \cl\ strategies: Elastic Weight Consolidation (\ewc), Memory Aware Synapses (\mas), and Learning without Forgetting (\lwf), chosen for their suitability in no-replay, privacy-conscious scenarios. Performance is analyzed using Word Error Rate (WER) for both \rnnt\ and \ctc\ paths on clean/noisy data, and knowledge retention via Backward Transfer. We explore varying training epochs (1, 2, 5 and 10) per task. Results, compared against naive fine-tuning, demonstrate \cl's efficacy in mitigating forgetting for scalable \asr\ in diverse Indian languages under realistic constraints. The code is available at \href{https://github.com/FrozenWolf-Cyber/Indic-CL-ASR}{https://github.com/FrozenWolf-Cyber/Indic-CL-ASR}
\end{abstract}

\section{Introduction}
India's extensive linguistic diversity poses significant hurdles for developing comprehensive Automatic Speech Recognition (\asr) systems \cite{zhong2024opportunities}. Traditional multilingual models, typically trained on aggregated datasets \cite{bai2021jointunsupervisedsupervisedtraining}, are ill-suited for real-world scenarios characterized by incremental data availability for low-resource languages, high computational costs of retraining, and data privacy concerns \cite{della2024cl}. Continual Learning (\cl), or lifelong learning \cite{ring1997child, de2021continual}, offers a paradigm to address these issues by enabling models to learn new tasks (languages) sequentially while preserving previously acquired knowledge. The primary challenge in \cl\ is catastrophic forgetting: the tendency of models to lose performance on past tasks when trained on new ones \cite{MCCLOSKEY1989109}. Mitigating this is crucial for successful \cl\ application \cite{Kirkpatrick_2017, goodfellow2015empiricalinvestigationcatastrophicforgetting}. This work applies \cl\ to multilingual \asr\ for Indian languages using the subset of the \indicSUPERB\ benchmark \cite{jain2024bhasaanuvaad}. We start with the \texttt{indicconformer} model (a Conformer-based \cite{gulati2020conformerconvolutionaugmentedtransformerspeech} hybrid \rnnt-\ctc\ \cite{burchi2024multilingual, graves2012sequence, graves2006connectionist} system pretrained on Hindi using NeMo \cite{Harper_NeMo_a_toolkit}) and incrementally train it on nine additional Indian languages: Bengali, Marathi, Telugu, Tamil, Urdu, Gujarati, Kannada, and Odia. We investigate three established \cl\ strategies: Elastic Weight Consolidation (\ewc) \cite{aich2021elastic}, Memory Aware Synapses (\mas) \cite{aljundi2018memory}, and Learning without Forgetting (\lwf) \cite{li2017learningforgetting}. These regularization and distillation methods are chosen because architecture-based approaches can inflate model size, and memory-based methods often violate realistic no-replay and privacy constraints \cite{rebuffi2017icarlincrementalclassifierrepresentation, lopezpaz2022gradientepisodicmemorycontinual}. Our experiments evaluate WER on clean and noisy data for both \rnnt\ and \ctc\ paths, and Backward Transfer to quantify forgetting, also varying training epochs per language. In summary, our contributions include: (1) the first comprehensive study of \cl\ for \asr\ across diverse Indian languages (2) systematic evaluation of \ewc, \mas, and \lwf\ under realistic constraints, and (3) detailed analysis of WER and knowledge retention across training regimes to guide practical deployment.

\section{Related Work}

Continual Learning (\cl) aims to enable AI systems to learn incrementally from a sequence of tasks without catastrophically forgetting prior knowledge. Key approaches include \cite{wang2024comprehensive} regularization-based methods (e.g., \ewc, which penalizes changes to parameters important for past tasks based on the Fisher Information Matrix; \mas, which uses the gradient of the squared L2 norm \cite{a92f3c16-7c6e-31d3-b403-82d2b0a469e4} of the model's output; SI \cite{zenke2017continual}), rehearsal-based methods (replaying past data) \cite{chaudhry2019efficientlifelonglearningagem}, and architecture-based methods (dynamically modifying model structure). Applying \cl\ to Automatic Speech Recognition (\asr) is challenging due to sequence variability, acoustic diversity, and linguistic complexity, especially when sequentially learning new languages in low-resource settings, common for many Indian languages. Hybrid \ctc-\rnnt\ models \cite{hori2017advancesjointctcattentionbased}, prevalent in modern \asr, offer multiple avenues for \cl\ integration. Our work explores \ewc, \mas, and (\lwf), which employs knowledge distillation to preserve the previous model's outputs on new data without storing old data. We utilize the subset of the \indicSUPERB\ benchmark \cite{jain2024bhasaanuvaad}, which provides standardized speech datasets for multiple Indian languages (including clean/noisy splits), and the \indicconformer, a state-of-the-art Conformer-based hybrid \rnnt-\ctc model pretrained on Hindi, as our base model and evaluation framework.

\vspace{-0.5em}
\section{Benchmark Design}
Our benchmark simulates realistic constraints for continual learning in multilingual \asr\ using the subset IndicSUPERB dataset. Each Indian language is treated as a separate task, forming a sequence of nine tasks beginning with Hindi ($T_1$), followed by Bengali, Marathi, Telugu, Tamil, Urdu, Gujarati, Kannada, and Odia ($T_2$ to $T_9$). All tasks are presented in a low-resource setting, with only 3000 training utterances per language (2000 clean and 1000 noisy). The model is trained sequentially using only the current task’s data $D_k$, enforcing a strict no-data-replay constraint. Training, validation, and test sets contain both clean and noisy samples, with test sets comprising 200 clean and 200 noisy utterances per language. Word Error Rate (WER) is evaluated separately on clean and noisy test splits using both \rnnt\ and \ctc\ decoding paths. To explore the trade-off between adaptation speed, accuracy on new tasks, and knowledge retention, we experiment with 1, 2, 5, and 10 training epochs per task. We benchmark performance against a naive sequential fine-tuning baseline. Further details on task formulation, model architecture, dataset construction and experimentation setup are provided in Appendix~\ref{sec:problem-formulation},  Appendix~\ref{sec:models}, Appendix~\ref{sec:dataset} and Appendix~\ref{sec:setup}.

\section{Evaluation Metrics}

\begin{itemize}
    \item \textbf{Word Error Rate (WER):} A commonly used metric in automatic speech recognition \cite{GOLDWATER2010181} and is expressed as a decimal fraction ranging from 0 to 1. WER is evaluated on all previously learned tasks after each new task is completed. Lower WER indicates better performance.

\item \textbf{Average Performance:} After training on task $T_k$, the average WER across all tasks $T_1$ to $T_k$ is given by:
\[
\text{AvgWER}_k = \frac{1}{k} \sum_{i=1}^{k} W_{k,i}
\]
where $W_{k,i}$ denotes the WER on task $T_i$ after learning task $T_k$. Lower AvgWER indicates better overall retention and adaptation.

\item \textbf{Backward Transfer (BWT):} Quantifies the influence of learning new tasks on the performance of previously learned ones. After task $T_k$, BWT is defined as:
\[
\text{BWT}_k = \frac{1}{k-1} \sum_{i=1}^{k-1} \left( \text{Acc}_{k,i} - \text{Acc}_{i,i} \right)
\]
where $\text{Acc}_{k,i} = 1 - W_{k,i}$ is the accuracy on task $T_i$ after learning task $T_k$, and $\text{Acc}_{i,i} = 1 - W_{i,i}$ is the accuracy on task $T_i$ immediately after it was learned. Higher BWT indicates better retention and less forgetting.

\end{itemize}

\begin{figure*}[ht]
\begin{adjustwidth}{-10cm}{-10cm}
\centering
\begin{subfigure}{0.6\textwidth}
\includegraphics[width=\linewidth]{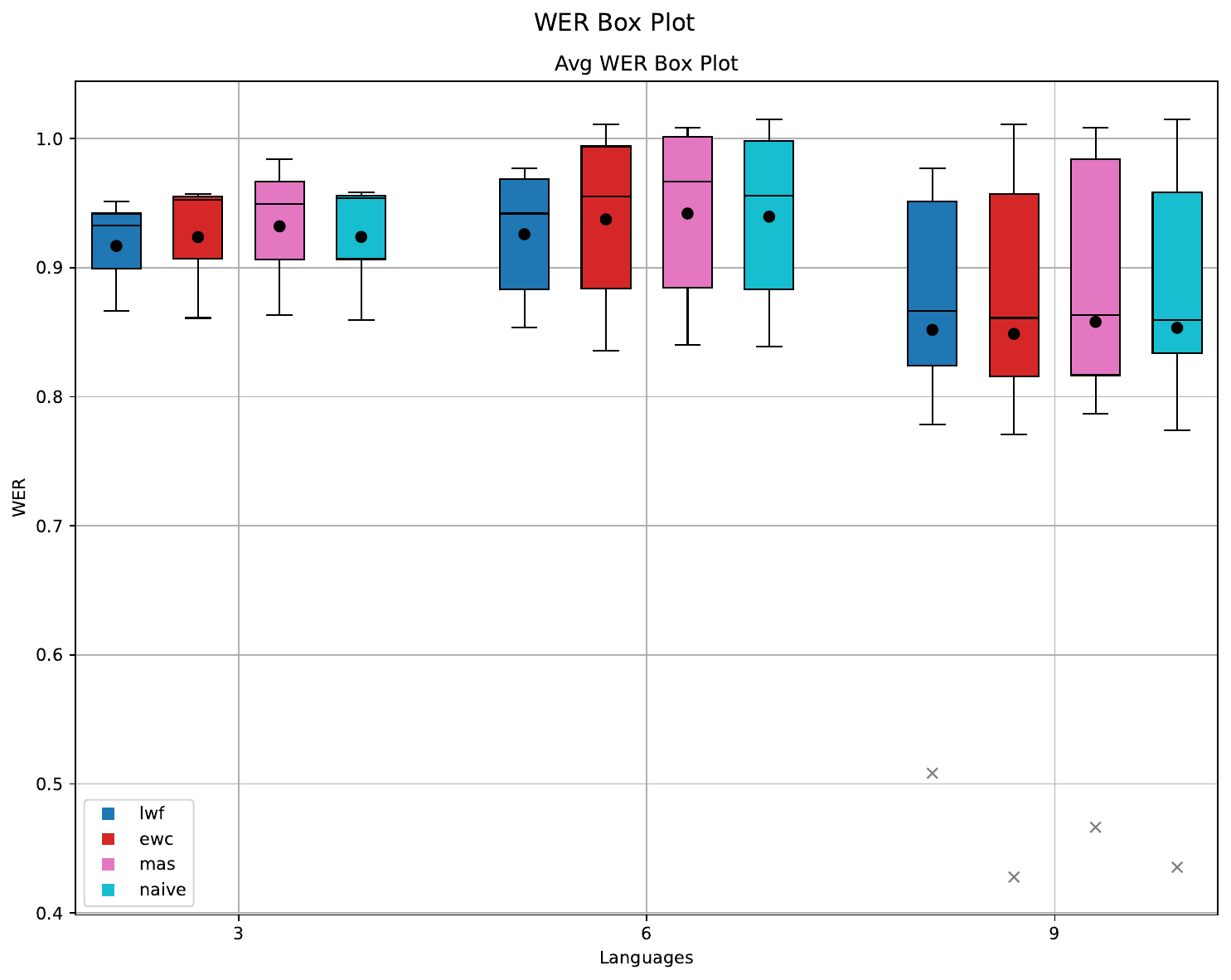}
\caption{CTC Box}
\end{subfigure}
\begin{subfigure}{0.6\textwidth}
\includegraphics[width=\linewidth]{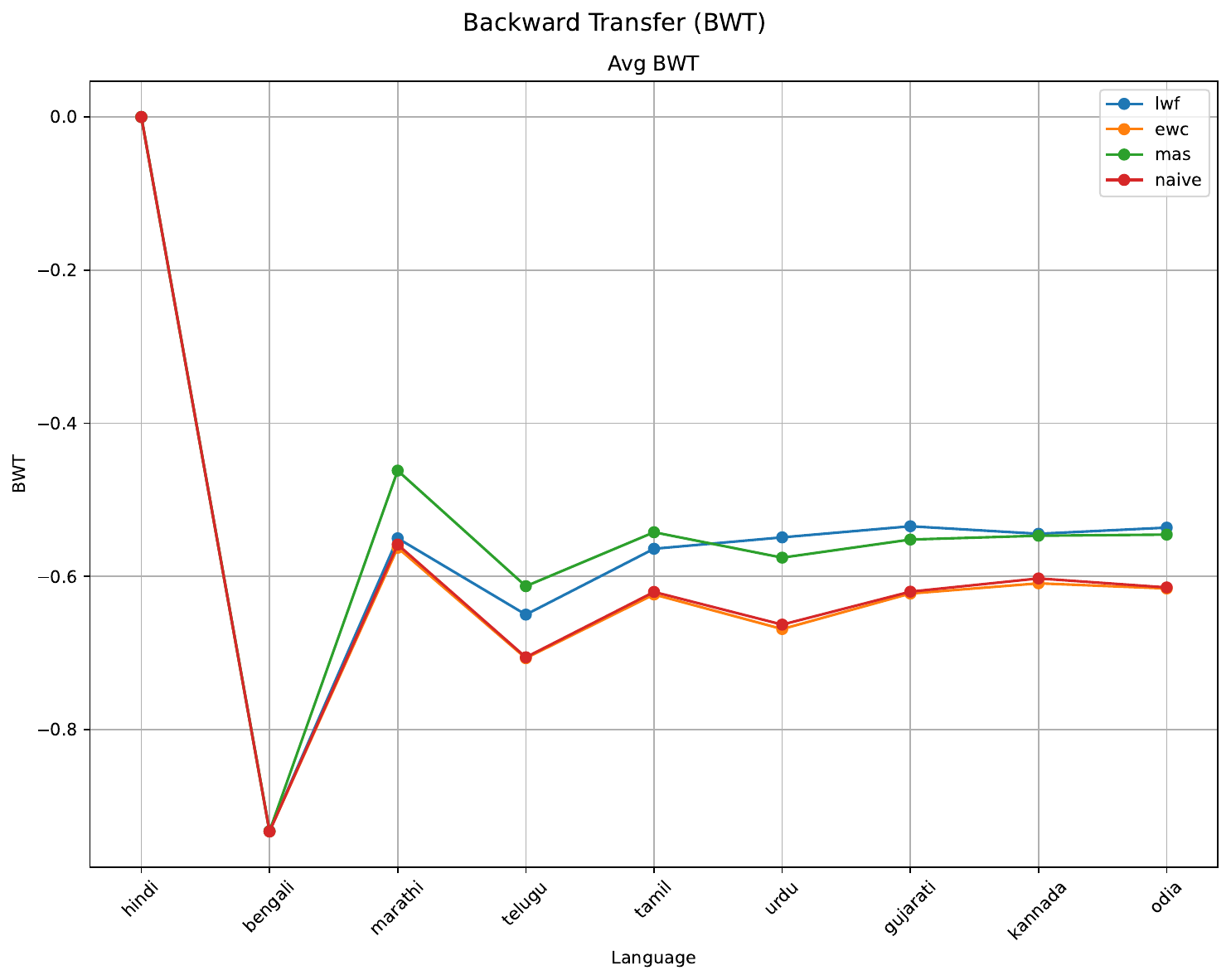}
\caption{CTC BWT}
\end{subfigure}
\caption{CTC Benchmark – Box and BWT Plots.}
\label{fig:ctc_benchmark_box_bwt}
\end{adjustwidth}
\end{figure*}
\section{Experiments and Results}

\subsection{Observations}

\paragraph{CTC Benchmarking}

As shown in \Cref{fig:ctc_benchmark_box_bwt}, the average WER across tasks reveals a clear ranking among methods. \textsc{LwF} achieves the best overall performance, followed by \textsc{EWC}, then \textsc{MAS}, with naive fine-tuning performing the worst. This ranking is particularly evident in short and medium task horizons. For longer sequences, however, the performance gap between methods narrows considerably. Naive fine-tuning, in particular, produces the highest WER maxima across tasks. When analyzing backward transfer (BWT), \textsc{MAS} performs best in short sequences, while \textsc{LwF} excels in medium-length tasks. For longer sequences, both \textsc{MAS} and \textsc{LwF} converge to similar average BWT values, whereas \textsc{EWC} and naive fine-tuning fall behind.

\paragraph{RNN-T Benchmarking}

\Cref{fig:rnnt_benchmark_box_bwt} shows that RNN-T \cite{xu2024multiblanktransducersspeechrecognition} consistently outperforms CTC in WER across all continual learning strategies. Among these, \textsc{EWC} achieves the lowest WER across task lengths, demonstrating strong performance retention on the current task. However, this benefit comes at a cost: \textsc{EWC} exhibits the worst BWT of all methods, even lower than that of naive fine-tuning, indicating substantial forgetting. \textsc{MAS} shows some improvement in BWT for medium-length sequences, but for longer horizons, BWT scores deteriorate across all methods except \textsc{EWC}, eventually becoming nearly indistinguishable.

\paragraph{General Comparison of CL Methods under Noisy Settings}

In noisy conditions (\Cref{fig:all_comparison_noisy_bwt_plot}), both \textsc{LwF} and \textsc{MAS} outperform \textsc{EWC} and the naive baseline in BWT, suggesting better retention of prior knowledge. Interestingly, noise appears to improve backward transfer, likely due to regularization effects. However, this improvement comes with a trade-off: WER increases, and models perform better on clean audio in absolute terms. This contrast indicates that noise can enhance stability, by reducing forgetting, while simultaneously impairing plasticity, by diminishing learning precision, which is reflected in the higher WER.

\paragraph{WER Performance Analysis}

\Cref{fig:all_comparison_noisy_wer_box_plot,fig:all_comparison_noisy_wer_shaded_plot} present WER trends over increasing task lengths. Evaluations are averaged over the last two and current tasks, categorized as short (1–3), medium (1–6), and long (1–9). In general, models perform better with clean data. Among the methods, \textsc{LwF} consistently maintains WER below 1.0, with high stability indicated by narrow shaded variance regions.

Interestingly, the upper bounds of noisy WER for \textsc{LwF} are comparable to the maxima seen under clean conditions. This can be attributed to its distillation-based loss, which prevents overfitting to noisy inputs by anchoring the model to previous predictions. \textsc{MAS} follows a similar pattern, though with slightly lower stability. \textsc{EWC} occasionally achieves better minimum WERs, particularly for short tasks, but continues to show poor BWT. The naive method performs surprisingly well in short sequences but fails to retain knowledge over longer horizons. Overall, \textsc{LwF} demonstrates the effectiveness of knowledge distillation in maintaining a balance between acquiring new knowledge and retaining previous learning. For longer sequences, average WER tends to decline, possibly due to simpler language characteristics in later tasks.

\paragraph{EWC Ablation Studies}

In \Cref{fig:ewc_ablation_box_shaded}, we examine the impact of different regularization strengths in \textsc{EWC} by testing $\lambda_{\text{EWC}} \in {5, 10}$. While both values yield similar outcomes, $\lambda_{\text{EWC}} = 10$ leads to slightly better WER in medium and long tasks, though the benefit is minimal in short tasks. BWT trends (\Cref{fig:bwt_all_ablation}) for both values remain close to those of the naive baseline, suggesting limited ability to retain performance on earlier tasks. Additionally, results from epoch-wise ablation (\Cref{fig:ewc_epoch_box_shaded}) show that increasing training epochs reduces WER, with the best results achieved at epoch 10. However, BWT steadily declines with more epochs (\Cref{fig:bwt_epoch_grid}), confirming the stability-plasticity trade-off: improved learning on new tasks often leads to increased forgetting of previous ones.
\vspace{-0.2em} 
\paragraph{LwF Ablation Studies}
\vspace{-0.2em} 
As shown in \Cref{fig:lwf_ablation_box_shaded}, adjusting the distillation weight (\(\alpha_{\text{KD}}\)) significantly impacts \textsc{LwF}’s performance. A higher value of 0.5 severely limits the model’s ability to learn new tasks, resulting in WERs close to 1.0 across all horizons thus worse than naive fine-tuning for short sequences. In contrast, \(\alpha_{\text{KD}} = 0.1\) strikes a better balance, achieving WER comparable to or better than naive fine-tuning while maintaining much stronger BWT. As shown in \Cref{fig:bwt_all_ablation}, the 0.5 configuration yields the highest BWT, primarily because the model barely updates and effectively freezes previous knowledge. The 0.1 setting enables more meaningful learning while controlling forgetting.

Epoch-wise trends (\Cref{fig:lwf_epoch_box_shaded,fig:bwt_epoch_grid}) are consistent with those observed in \textsc{EWC}. Increasing the epochs improves WER but worsens BWT.
\vspace{-0.2em} 
\paragraph{MAS Ablation Studies}
\vspace{-0.2em} 
In \Cref{fig:mas_ablation_box_shaded}, we compare \textsc{MAS} with regularization weights \(\alpha_{\text{ctx}}\) of 0.3 and 1.0. The stronger setting of 1.0 consistently achieves better WER and shows more stable variance across tasks. Its shaded performance region closely overlaps with that of naive fine-tuning, though with lower dispersion. When examining BWT (\Cref{fig:bwt_all_ablation}), the 0.3 configuration performs better, matching \textsc{LwF} in retaining knowledge.

As with the other methods, \textsc{MAS} exhibits the stability-plasticity trade-off: increasing epochs (\Cref{fig:mas_epoch_box_shaded}) lowers WER but leads to worsening BWT (\Cref{fig:bwt_epoch_grid}). This consistent trend across methods emphasizes the fundamental challenge in continual learning of effectively balancing the acquisition of new information with the retention of existing knowledge.

\vspace{-0.5em} 
\section{Discussion}
\vspace{-0.5em} 
Our findings show that LwF and MAS generally offer better BWT in noisy ASR, indicating superior retention of prior languages. The inverse link between noise-driven BWT improvement and WER degradation suggests noise acts as an implicit regularizer, improving retention at the cost of transcription accuracy. LwF’s consistently low and stable WER, especially in longer task sequences, highlights its distillation-based regularization effectiveness in noisy settings by preventing over-adaptation. In contrast, EWC, while competitive in shorter tasks or with RNN-T, often showed poor BWT, particularly with RNN-T, indicating weight consolidation is less effective for complex recurrent models or sequential multilingual learning.

Ablation studies confirmed the stability-plasticity dilemma. Longer training improves current task WER but worsens BWT. Stronger regularization improves BWT but hinders new learning, while weaker regularization enhances plasticity but increases forgetting. Comparing CTC and RNN-T, RNN-T achieved better WER but worsened catastrophic forgetting, especially for EWC. The decline of BWT in long RNN-T sequences, except for EWC, highlights challenges for current CL methods with advanced ASR models over extended tasks. Notably, despite CL, absolute WER during new task learning remains suboptimal for practical use, underscoring the difficulty in balancing plasticity and retention and the early stage of CL in ASR.
\vspace{-0.7em} 
\section{Conclusion}
\vspace{-0.7em} 
This study shows that while LwF and MAS can improve BWT in noisy, multi-language ASR compared to baselines and EWC, a fundamental trade-off persists. Noise appears to aid BWT, possibly as a regularizer, but consistently degrades WER. LwF offered the most balanced performance with stable, low WER and good BWT for longer sequences. The stability-plasticity dilemma was pervasive: efforts to improve new task learning typically increased forgetting. RNN-T models, while delivering superior WER, amplified catastrophic forgetting. Importantly, even with CL, overall WER during new language learning often remains too high for practical deployment. This signals that current CL methods are not yet complete solutions and that CL in ASR requires further investigation for real-world viability.

\section{Limitations}

While our work offers valuable insights into continual learning (CL) for multilingual ASR under noise, several limitations must be acknowledged. First, the study does not systematically investigate the impact of language ordering on performance. Since language sequence can significantly influence both task difficulty and forgetting dynamics, this is a key variable requiring further exploration. Second, our findings are constrained to the specific datasets, noise profiles, and ASR architectures (CTC and RNN-T) evaluated. As such, the extent to which these results generalize to other languages, domains, or ASR models (e.g., Transformer-based architectures) remains uncertain.

\section{Future Work}

To advance CL for ASR towards practical applications, future work should explore:
\begin{itemize}
    \item \textbf{Federated learning frameworks} \cite{Bharati_2022} to address privacy and simulate realistic distributed ASR deployment.
    \item Transitioning to \textbf{online learning paradigms} where data arrives as a continuous stream, reflecting many real-world ASR use-cases and posing new challenges for CL algorithm efficiency \cite{harun2023efficienttodayscontinuallearning} and adaptability.
    \item The resilience and adaptation of CL strategies in \textbf{adversarial settings} \cite{ebrahimi2020adversarialcontinuallearning} to develop more secure and reliable systems.
    \item Developing \textbf{novel CL techniques} specifically tailored to speech's sequential nature and modern ASR model intricacies (e.g., RNN-T) to better overcome the stability-plasticity dilemma and achieve deployment-ready performance.
\end{itemize}

\section{Acknowledgment}
This work was supported by the IndiaAI Fellowship, which provided the financial assistance necessary to undertake this research. Computational experiments were conducted using the PARAM Porul supercomputing facility at the National Institute of Technology, Tiruchirappalli, under the National Supercomputing Mission (NSM). The facility is implemented by the Centre for Development of Advanced Computing (C-DAC) and supported by the Ministry of Electronics and Information Technology (MeitY) and the Department of Science and Technology (DST), Government of India.

\bibliography{custom}  




\appendix
\section{Appendix}

\subsection{Problem Formulation}
\label{sec:problem-formulation}

We formulate the continual learning (CL) problem in multilingual \asr\ as a sequential learning setup. Let $D = \{D_1, D_2, ..., D_N\}$ denote a sequence of datasets, each corresponding to a task $T_k$ (i.e., language $k$). Each dataset $D_k = \{(x_{kj}, y_{kj})\}$ contains speech utterances $x_{kj}$ and transcriptions $y_{kj}$. The goal is to train an \asr\ model $M(\theta)$ over tasks $T_1, ..., T_N$ such that it learns the current task well while preserving performance on previous tasks.

During training on task $T_k$, only data $D_k$ is accessible. A naive fine-tuning approach minimizes the loss for task $T_k$ starting from the parameters $\theta_{k-1}$ obtained from the previous task:

\[
\theta_k = \arg\min_{\theta} L_k(\theta),
\]

where $L_k(\theta)$ is the task-specific loss composed of a weighted sum of \rnnt\ and \ctc\ objectives. However, such fine-tuning often causes \emph{catastrophic forgetting}, where performance degrades significantly on previously learned tasks.

To address this, we integrate three regularization-based CL methods into our training pipeline:

\begin{itemize}
    \item \textbf{Elastic Weight Consolidation (EWC):} Prevents drift on important parameters by adding a quadratic penalty based on a Fisher Information matrix estimated after each task. The updated loss becomes:
    \[
    L_{\text{total}} = L_k(\theta) + \lambda_{\text{EWC}} \sum_j F_j (\theta_j - \theta_j^*)^2,
    \]
    where $F_j$ is the accumulated Fisher importance and $\theta^*$ are parameters from the previous task.
    
    \item \textbf{Memory Aware Synapses (MAS):} Estimates importance via gradients of the squared norm of model outputs (logits) and adds a similar penalty:
    \[
    L_{\text{total}} = L_k(\theta) + \lambda_{\text{MAS}} \sum_j \Omega_j (\theta_j - \theta_j^*)^2,
    \]
    where $\Omega_j$ is the importance computed from absolute gradients w.r.t. combined \rnnt\ and \ctc\ output activations.
    
    \item \textbf{Learning without Forgetting (LwF):} Adds a distillation loss to encourage the current model to produce similar outputs as the frozen model from the previous task:
    \[
    L_{\text{total}} = (1 - \alpha) \cdot L_k(\theta) + \alpha \cdot L_{\text{distill}},
    \]
    where $L_{\text{distill}}$ is a weighted combination of KL divergence or MSE between the current and previous model’s \rnnt\ and \ctc\ outputs on current task data.
\end{itemize}

In our setup:
\begin{itemize}
    \item Tasks $T_1 \ldots T_{9}$ correspond to the 9 Indian languages in \indicSUPERB.
    \item Only $D_k$ is available while training on task $T_k$.
    \item The model $M(\theta_0)$ is initialized from a Hindi-pretrained \indicconformer.
    \item The base loss $L_k$ is:
    \[
    L_k(\theta) = (1 - w_{\text{CTC}}) \cdot L_{\text{RNNT}} + w_{\text{CTC}} \cdot L_{\text{CTC}}.
    \]
\end{itemize}

This formulation allows us to balance plasticity (learning new tasks) and stability (retaining performance on past tasks) through principled integration of CL techniques.

\subsection{Dataset}
\label{sec:dataset}

We conduct our experiments using the \indicSUPERB\ benchmark, which originally encompasses 11 Indian languages. For this study, we focus on nine languages: Hindi (hi), Bengali (bn), Marathi (mr), Telugu (te), Tamil (ta), Urdu (ur), Gujarati (gu), Kannada (kn), and Odia (or). These languages cover both the Indo-Aryan and Dravidian families, ensuring linguistic diversity.

To simulate a low-resource scenario, we utilize a subset of 3,000 training utterances per language, composed of 2,000 clean and 1,000 noisy samples. The validation and test sets each consist of 400 utterances, evenly split between clean and noisy conditions. This consistent setup allows us to rigorously evaluate model performance under constrained data conditions across multiple languages.
\subsection{Model Architecture}
\label{sec:models}

Our automatic speech recognition system (\indicconformer) is built around a hybrid architecture that combines a Conformer-based encoder with both Recurrent Neural Network Transducer (\rnnt) and Connectionist Temporal Classification (\ctc) objectives using NeMo \cite{Harper_NeMo_a_toolkit}. The Conformer encoder effectively captures speech features by integrating convolutional layers to model local dependencies alongside self-attention mechanisms for global context.

The \rnnt\ component models output sequences in an end-to-end fashion, composed of an encoder, a prediction network that autoregressively generates hypotheses based on previous tokens, and a joint network that fuses these signals. This structure inherently manages acoustic modeling and alignment without requiring explicit segmentation.

In parallel, the \ctc\ loss facilitates training without frame-level alignment by introducing a blank token and summing probabilities over all valid alignments. Often used as an auxiliary objective, \ctc\ guides the encoder towards robust and stable feature representations.

We train the model by jointly optimizing the \rnnt\ and \ctc\ losses, combining them in a weighted sum:
\[
L_{\text{base}} = (1 - w_{\text{CTC}}) \cdot L_{\text{RNNT}} + w_{\text{CTC}} \cdot L_{\text{CTC}}
\]
where \(w_{\text{CTC}}\) is the weight for the \ctc\ loss.

\subsection{Continual Learning Methods Implementation}

To mitigate forgetting in continual learning, we augment the base loss with regularization losses depending on the method used.

\subsubsection{Learning without Forgetting (LwF)}

LwF employs a knowledge distillation loss using KL-divergence \cite{kullback1951information} that encourages the current model to mimic the outputs of the frozen previous model on the new data. Distillation is applied separately on the \rnnt\ logits and \ctc\ output probabilities.

\begin{align*}
L_{\text{dist}}^{\text{RNNT}} &= \text{DistillationLoss}\big(O_{\text{RNNT}}(\theta),\, O_{\text{RNNT}}(\theta^*)\big), \\
L_{\text{dist}}^{\text{CTC}} &= \text{DistillationLoss}\big(O_{\text{CTC}}(\theta),\, O_{\text{CTC}}(\theta^*)\big),
\end{align*}

where \(O_{\text{RNNT}}\) and \(O_{\text{CTC}}\) denote the outputs (logits or probabilities) of the current and frozen models respectively.

The total distillation loss is a weighted sum:
\[
L_{\text{dist}} = (1 - \alpha_{\text{ctx}}) \cdot L_{\text{dist}}^{\text{RNNT}} + \alpha_{\text{ctx}} \cdot L_{\text{dist}}^{\text{CTC}},
\]
with \(\alpha_{\text{ctx}} \in [0,1]\) balancing between \rnnt\ and \ctc\ distillation.

Finally, the full training loss is:
\[
L_{\text{total}} = (1 - \alpha_{\text{KD}}) \cdot L_{\text{base}} + \alpha_{\text{KD}} \cdot L_{\text{dist}},
\]
where \(\alpha_{\text{KD}} \in [0,1]\) controls the strength of the knowledge distillation regularization.

\subsubsection{Memory Aware Synapses (MAS)}

MAS estimates parameter importance by measuring the sensitivity of the squared norm of the model's outputs to each parameter. This is done separately for the \ctc\ decoder and the \rnnt\ joint network logits.

First, compute the squared logit norms and average over the batch:

\[
L_{\text{CTC\_logits}} = \frac{1}{B} \sum_{b=1}^B \left\| \mathbf{z}_{\text{CTC}}^{(b)} \right\|_2^2
\]
where \(\mathbf{z}_{\text{CTC}}^{(b)}\) are the flattened \ctc\ decoder logits for batch element \(b\).

Similarly, compute the average squared norm over the stored \rnnt\ joint network logits:

\[
L_{\text{RNNT\_logits}} = \frac{1}{N} \sum_{n=1}^N \frac{1}{B} \sum_{b=1}^B \left\| \mathbf{z}_{\text{RNNT}, n}^{(b)} \right\|_2^2,
\]
where \(\mathbf{z}_{\text{RNNT}, n}^{(b)}\) is the flattened joint logits tensor stored at step \(n\), and \(N\) is the total number of stored logits.

Combine these with a weighting factor \(\alpha_{\text{ctx}} \in [0,1]\):

\[
L_{\text{logits}} = (1 - \alpha_{\text{ctx}}) \cdot L_{\text{RNNT\_logits}} + \alpha_{\text{ctx}} \cdot L_{\text{CTC\_logits}}.
\]

Perform backpropagation on \(L_{\text{logits}}\) to obtain gradients \(\nabla_{\theta_j} L_{\text{logits}}\). Then, update parameter importance values as the accumulated absolute gradients:

\[
\Omega_j \leftarrow \Omega_j + \left| \frac{\partial L_{\text{logits}}}{\partial \theta_j} \right|.
\]

Finally, the MAS regularization penalty is computed as:

\[
L_{\text{MAS}} = \lambda_{\text{MAS}} \sum_j \Omega_j (\theta_j - \theta_j^*)^2,
\]

where \(\lambda\) is the MAS regularization strength, and \(\theta_j^*\) are the parameters saved after the previous task.

The full training loss is:

\[
L_{\text{total}} = L_{\text{base}} + L_{\text{MAS}}.
\]

\subsubsection{Elastic Weight Consolidation (EWC)}

EWC mitigates catastrophic forgetting by penalizing changes to parameters deemed important for previously learned tasks. Importance is quantified using the diagonal of the Fisher Information Matrix.

After task \(T_i\), the diagonal Fisher is estimated as:

\[
F_{i,j} = \mathbb{E}_{x \sim D_i} \left[ \left( \frac{\partial L_i(\theta)}{\partial \theta_j} \right)^2 \right],
\]

where \(F_{i,j}\) denotes the importance of parameter \(\theta_j\) and is computed by averaging squared gradients over the dataset \(D_i\).

To accumulate importance across tasks, we update the consolidated Fisher with a decay factor \(\gamma\):

\[
F_{\text{consol}, i} = \gamma \cdot F_{\text{consol}, i-1} + F_{i}.
\]

This allows older tasks’ importance to gradually decay while emphasizing more recent tasks.

During training on a new task, the EWC penalty is added to the base loss:

\[
L_{\text{EWC}} = \lambda_{\text{EWC}} \sum_j F_{\text{consol}, j} (\theta_j - \theta_j^*)^2,
\]

where \(\theta_j^*\) are the parameter values saved after the previous task, and \(\lambda\) controls the regularization strength.

The full training loss becomes:

\[
L_{\text{total}} = L_{\text{base}} + L_{\text{EWC}}.
\]

In practice, the penalty gradient with respect to each parameter \(\theta_j\) is computed as:

\[
\frac{\partial L_{\text{EWC}}}{\partial \theta_j} = 2\lambda \cdot F_{\text{consol}, j} (\theta_j - \theta_j^*),
\]

which directly enters the optimization step during gradient update.

\subsubsection{Summary of Hyperparameters}

\begin{itemize}
  \item \(w_{\text{CTC}}\): Weight of \ctc\ loss in the base loss.
  \item \(\alpha_{\text{KD}}\): Weight of the knowledge distillation loss in LwF.
  \item \(\alpha_{\text{ctx}}\): Balancing weight between \rnnt\ and \ctc\ components in distillation and MAS.
  \item \(\lambda\): Regularization strength for MAS and EWC.
\end{itemize}

\subsection{Experimental Setup}
\label{sec:setup}

All experiments are conducted on an NVIDIA V100 GPU using the XXX supercomputer SLURM cluster. Each run took about 13 hours to 3 days depending on the ablation hyper parameters. We initialize our models with the \indicconformer\ pretrained on Hindi (\texttt{ai4bharat/indicconformer\_stt\_hi\_hybrid\_rnnt\_large}
) using NeMo \cite{Harper_NeMo_a_toolkit}, providing a strong starting point for multilingual speech recognition. The model used in our experiments consists of approximately 130 million parameters. The dataset consists of the \indicSUPERB\ benchmark split across nine Indian languages.

Our continual learning experiments follow a fixed sequence of tasks: Hindi $\rightarrow$ Bengali $\rightarrow$ Marathi $\rightarrow$ Telugu $\rightarrow$ Tamil $\rightarrow$ Urdu $\rightarrow$ Gujarati $\rightarrow$ Kannada $\rightarrow$ Odia. For each new task, the model is initialized from the previously trained model and trained exclusively on the current language’s data (3,000 samples: 2,000 clean and 1,000 noisy).

Training is performed for varying numbers of epochs (1, 2, 5, and 10) to evaluate how training duration impacts model performance and forgetting. Optimization is done using Adam \cite{kingma2014adam} with a learning rate of $1 \times 10^{-4}$.

We apply the following continual learning parameters:
\begin{itemize}
    \item \textbf{Elastic Weight Consolidation (EWC)} with $\lambda_{\text{MAS}} \in \{10, 5\}$ and $\gamma = 1.0$
    \item \textbf{Memory Aware Synapses (MAS)} with $\lambda_{\text{MAS}} = 1 $ and $\alpha_{\text{ctx}} \in \{0.3, 1.0\}$
    \item \textbf{Learning without Forgetting (LwF)} with $\alpha_{\text{KD}} \in \{0.1, 0.5\}$ and $\alpha_{\text{ctx}} = 0.3$
\end{itemize}

The base model is trained using a weighted combination of \rnnt\ and \ctc\ losses with weights:
\[
w_{\text{RNNT}} = 0.7, \quad w_{\text{CTC}} = 0.3
\]

As a baseline, naive fine-tuning (training on each new task without any continual learning strategy) is also evaluated.

After training on each task $T_k$, we evaluate the model on the test sets of all tasks from $T_1$ through $T_k$. This allows us to compute Word Error Rate (WER) and continual learning metrics such as average accuracy, forgetting, and retention. Hyperparameters and optimization settings are kept consistent across all methods and tasks to ensure fair and reproducible comparisons.

\begin{figure*}[ht]
\begin{adjustwidth}{-6cm}{-6cm}
\centering
\includegraphics[width=0.7\linewidth]{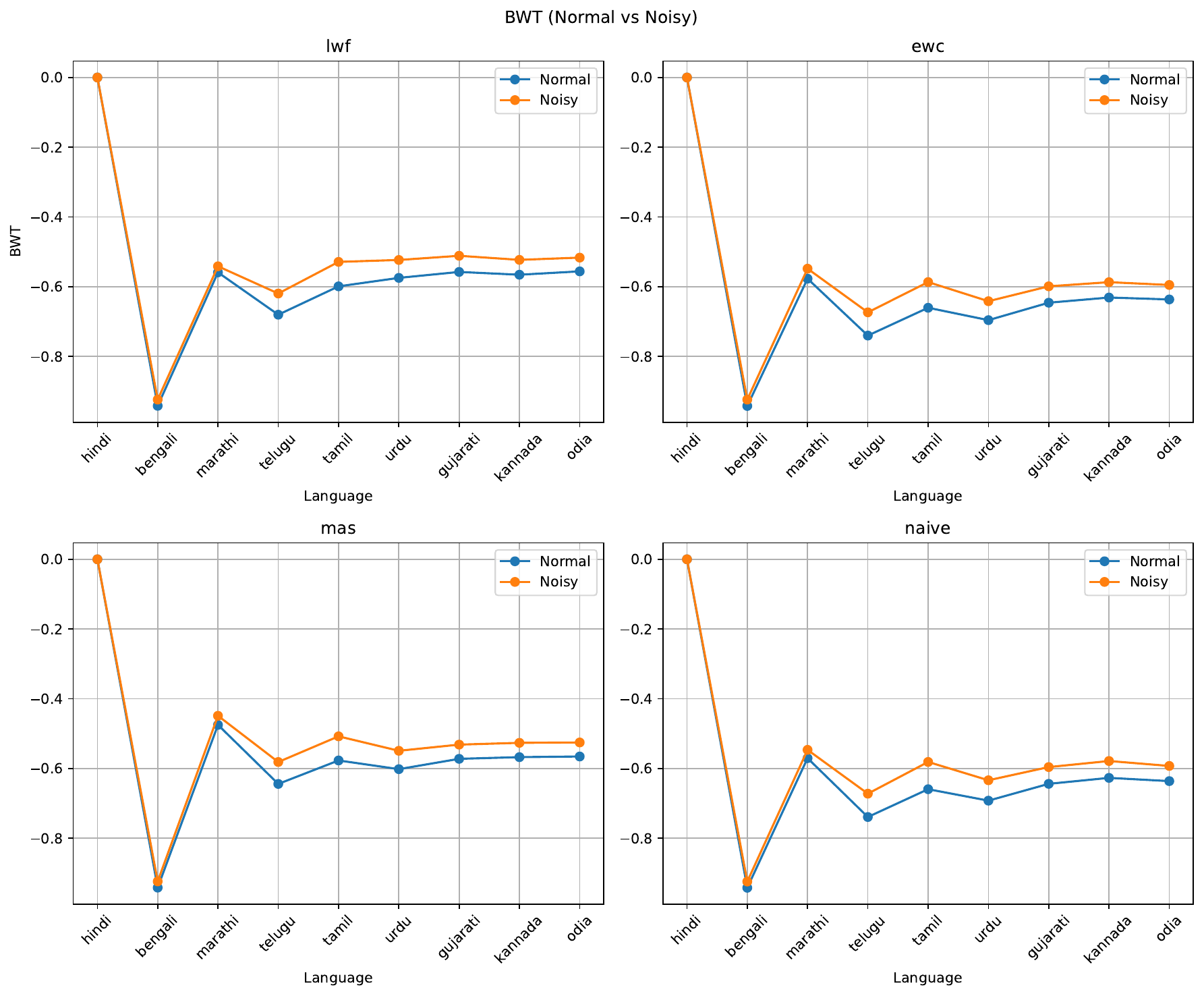}
\caption{All comparison noisy BWT plot}
\label{fig:all_comparison_noisy_bwt_plot}
\end{adjustwidth}
\end{figure*}

\begin{figure*}[ht]
\begin{adjustwidth}{-6cm}{-6cm}
\centering
\includegraphics[width=0.7\linewidth]{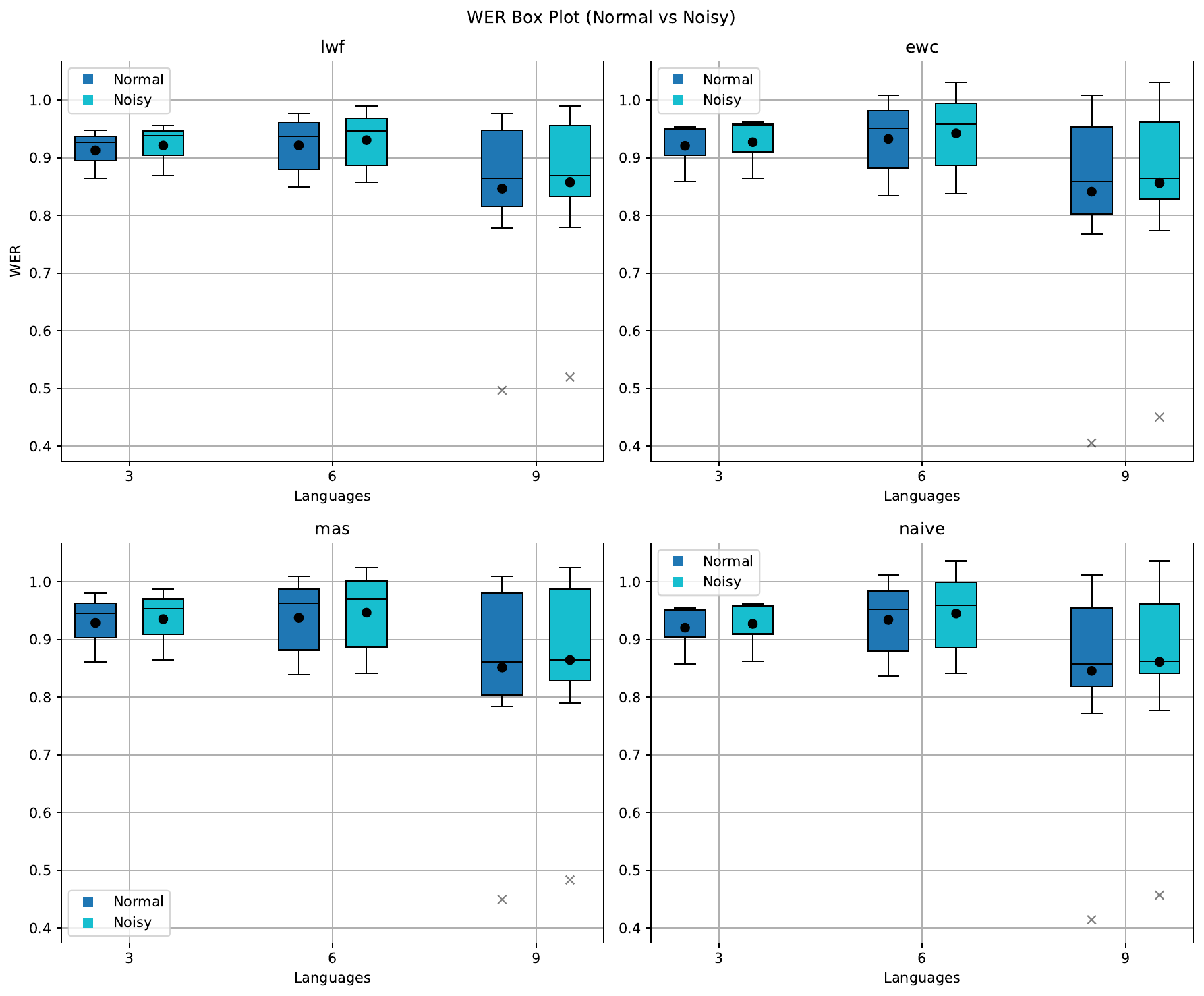}
\caption{All comparison noisy WER box plot}
\label{fig:all_comparison_noisy_wer_box_plot}
\end{adjustwidth}
\end{figure*}

\begin{figure*}[ht]
\begin{adjustwidth}{-6cm}{-6cm}
\centering
\includegraphics[width=0.7\linewidth]{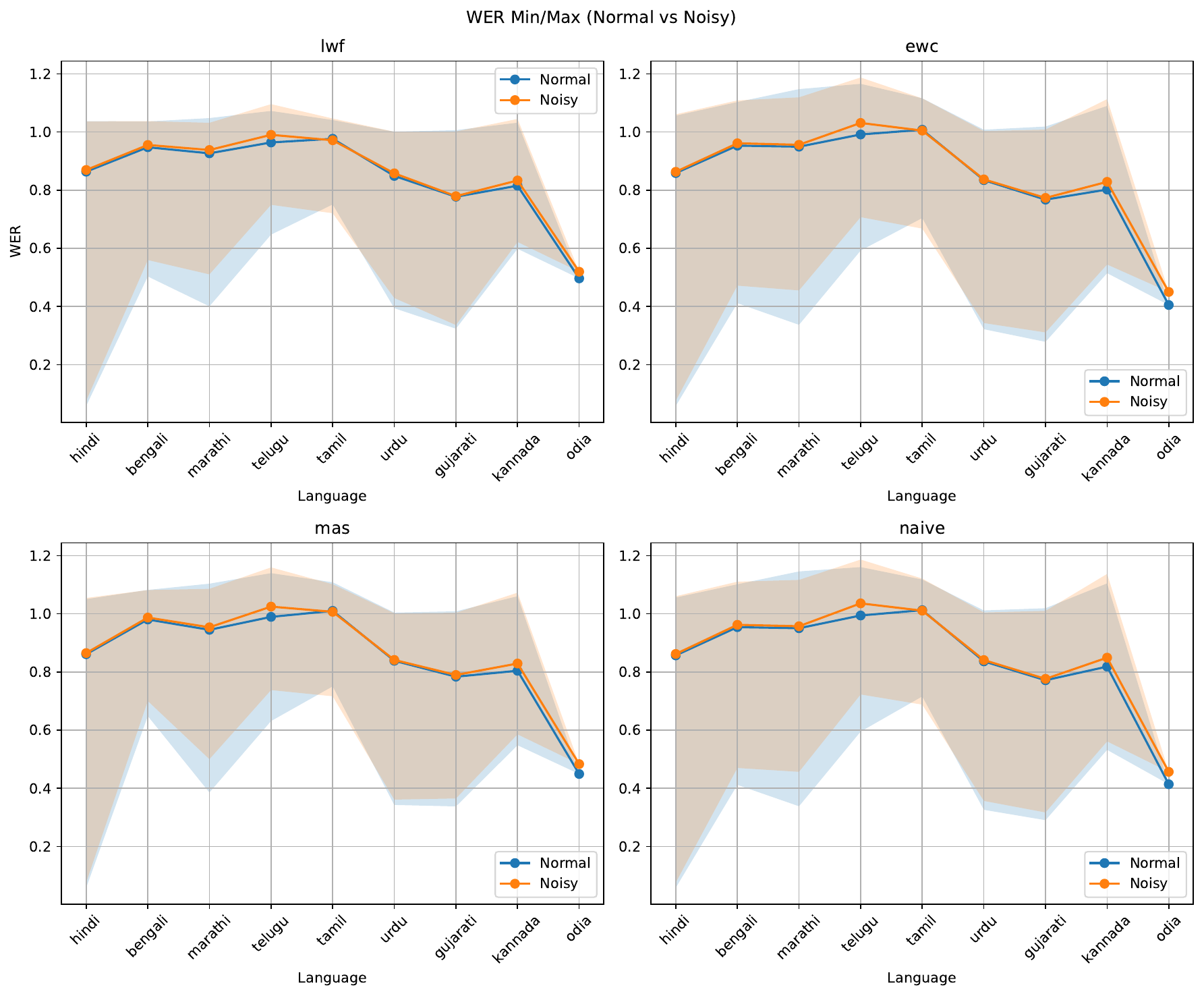}
\caption{All comparison noisy shaded WER plot}
\label{fig:all_comparison_noisy_wer_shaded_plot}
\end{adjustwidth}
\end{figure*}


\begin{figure*}[ht]
\begin{adjustwidth}{-10cm}{-10cm}
\centering
\begin{subfigure}{0.6\textwidth}
\includegraphics[width=\linewidth]{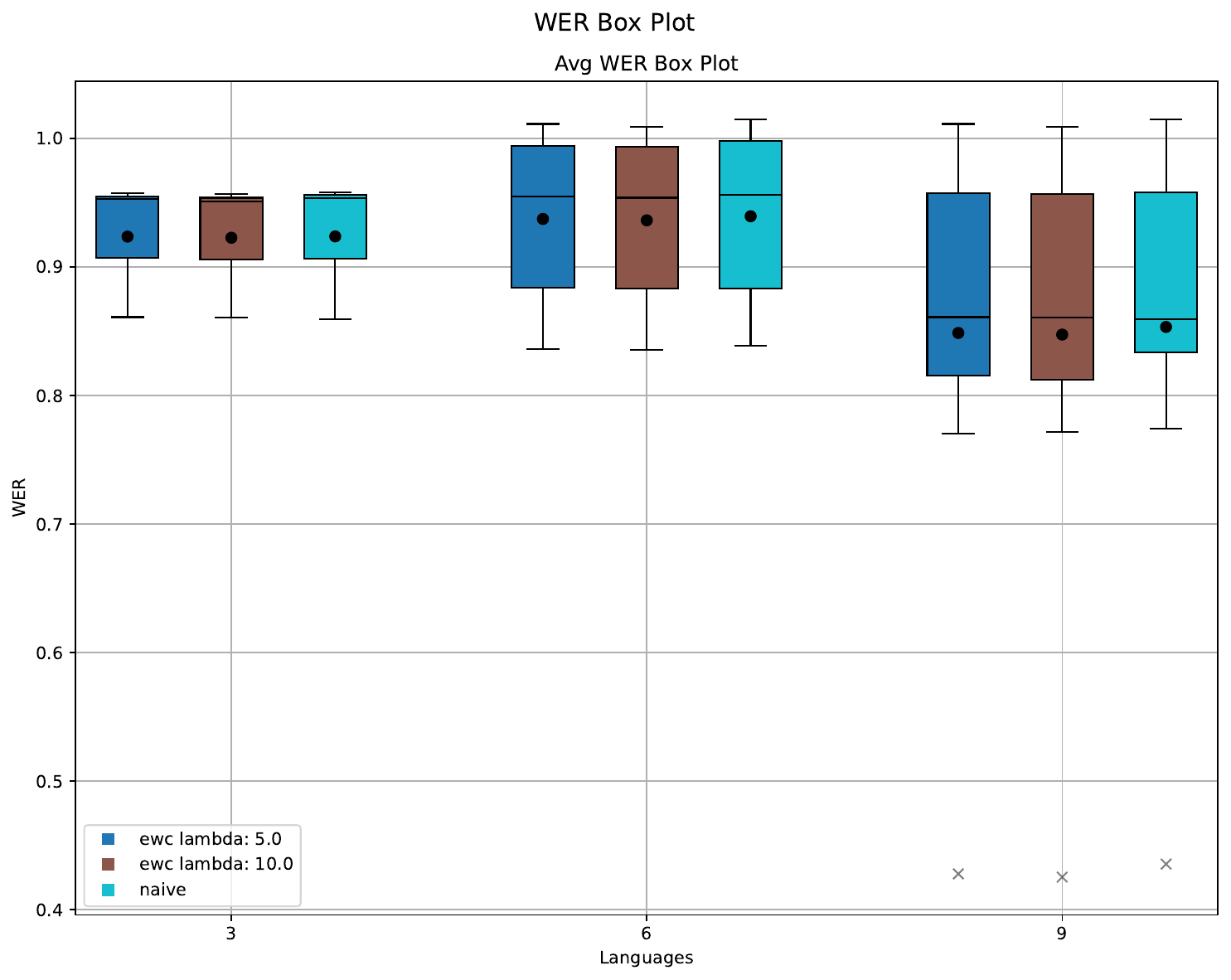}
\caption{EWC Box}
\end{subfigure}
\begin{subfigure}{0.6\textwidth}
\includegraphics[width=\linewidth]{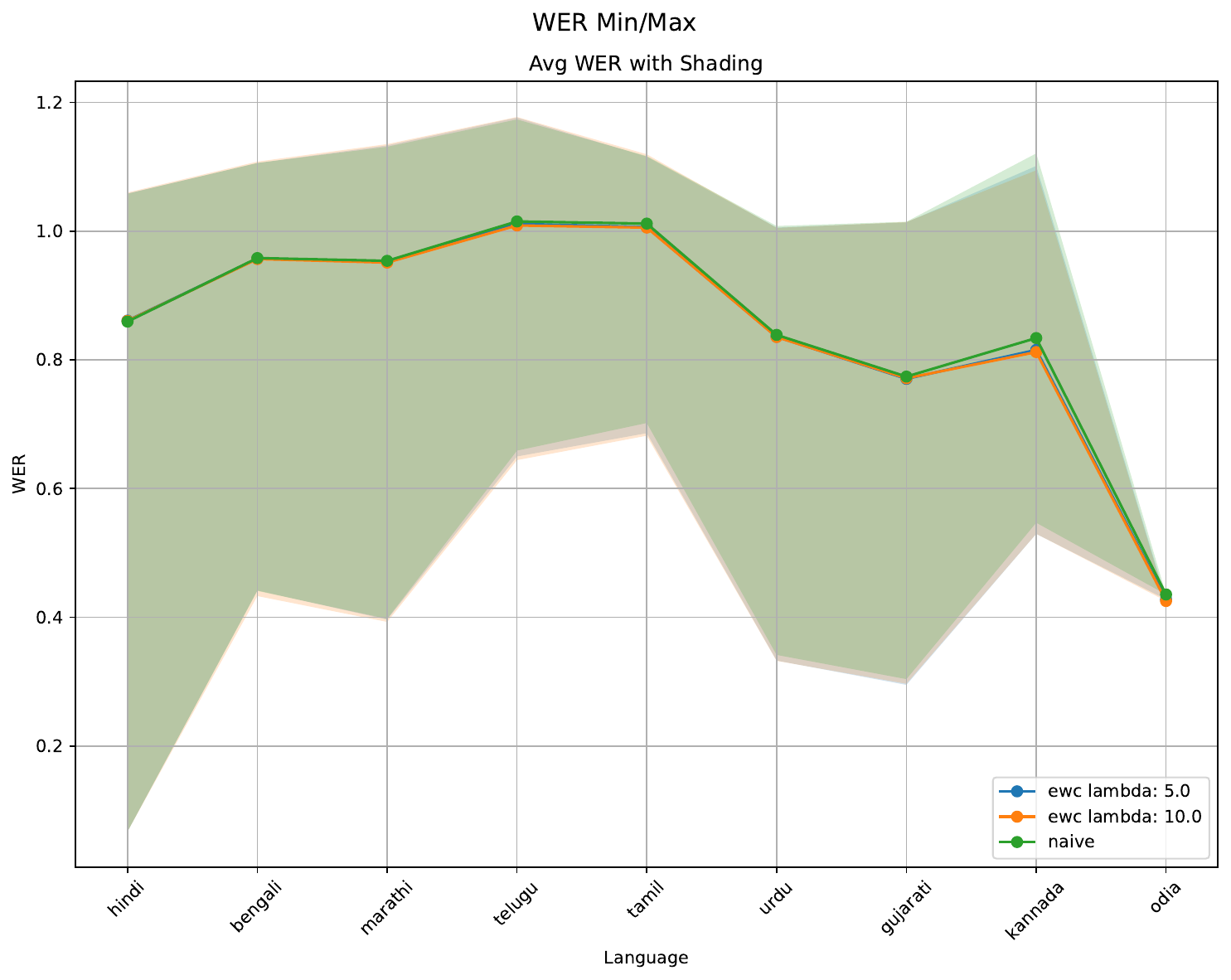}
\caption{EWC Shaded}
\end{subfigure}
\caption{EWC Ablation – Box and Shaded Plots}
\label{fig:ewc_ablation_box_shaded}
\end{adjustwidth}
\end{figure*}

\begin{figure*}[ht]
\begin{adjustwidth}{-10cm}{-10cm}
\centering
\begin{subfigure}{0.6\textwidth}
\includegraphics[width=\linewidth]{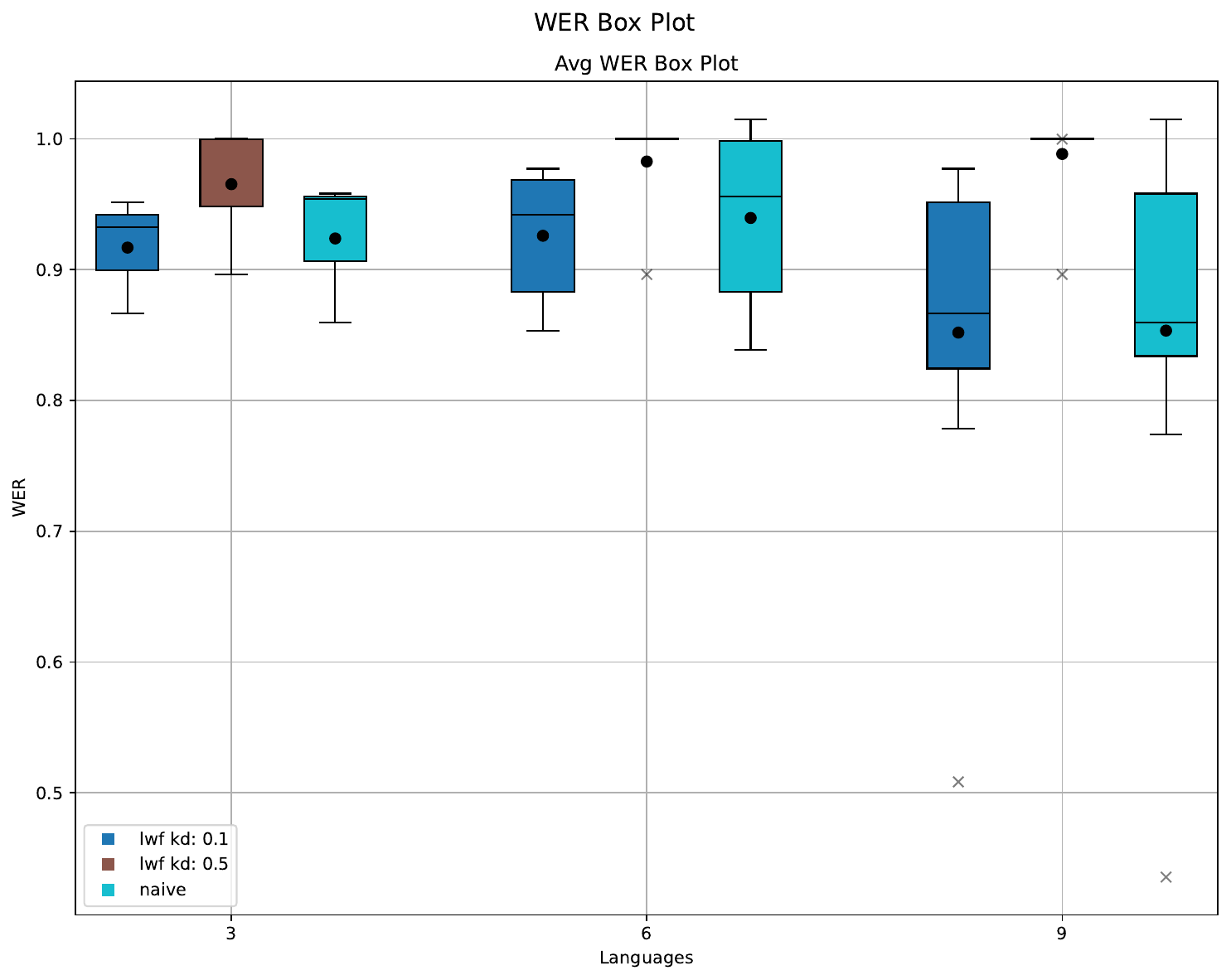}
\caption{LWF Box}
\end{subfigure}
\begin{subfigure}{0.6\textwidth}
\includegraphics[width=\linewidth]{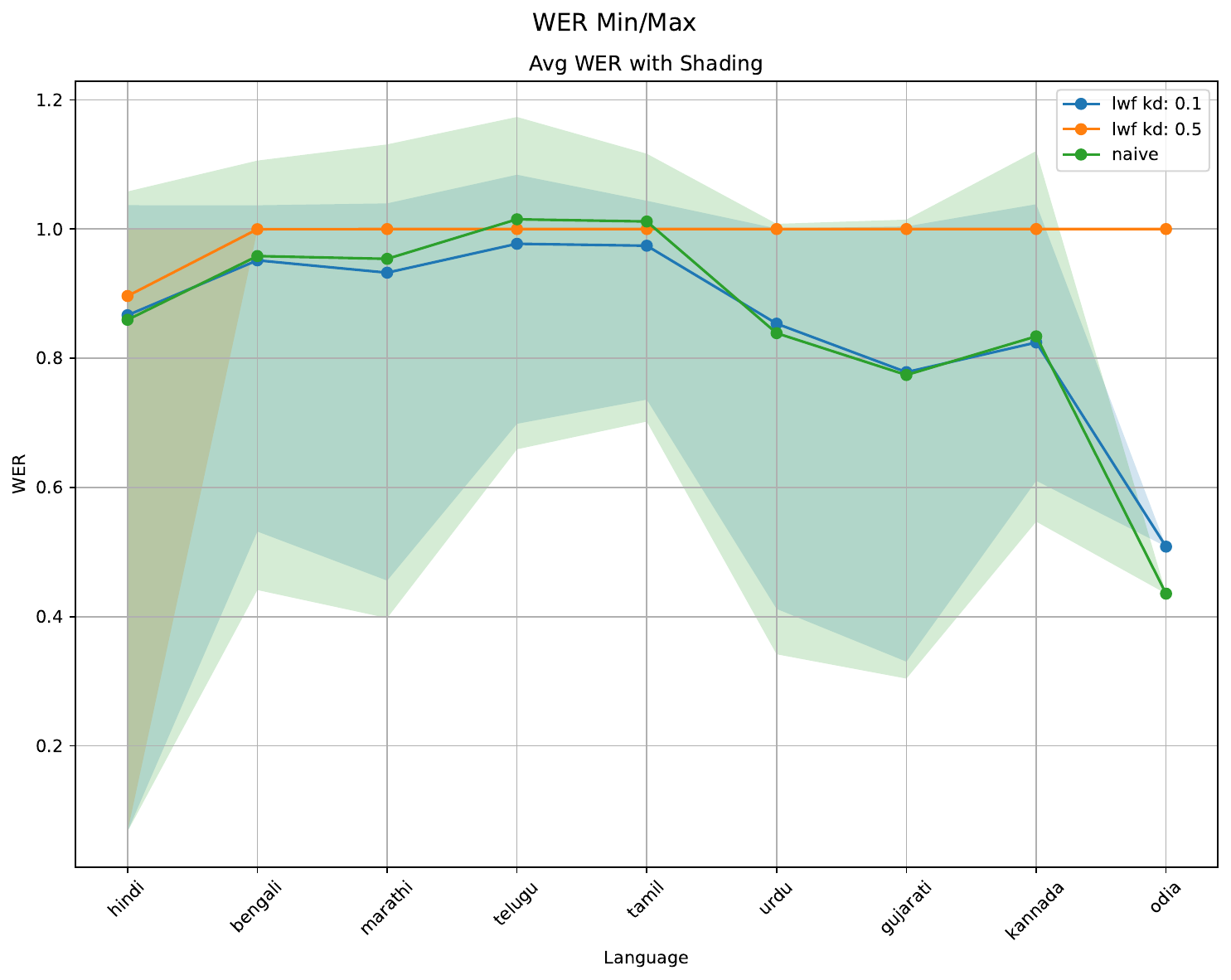}
\caption{LWF Shaded}
\end{subfigure}
\caption{LWF Ablation – Box and Shaded Plots}
\label{fig:lwf_ablation_box_shaded}
\end{adjustwidth}
\end{figure*}

\begin{figure*}[ht]
\begin{adjustwidth}{-10cm}{-10cm}
\centering
\begin{subfigure}{0.6\textwidth}
\includegraphics[width=\linewidth]{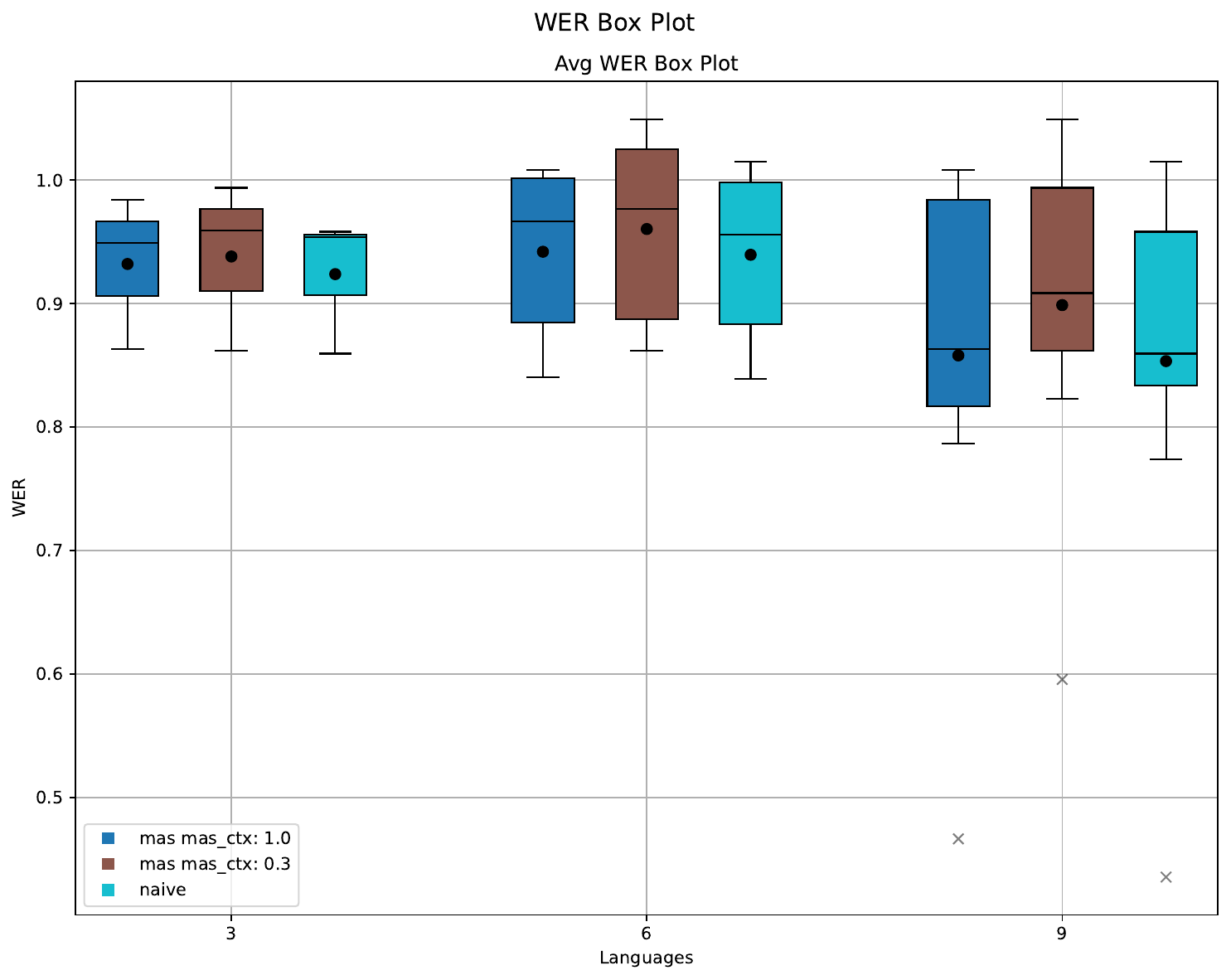}
\caption{MAS Box}
\end{subfigure}
\begin{subfigure}{0.6\textwidth}
\includegraphics[width=\linewidth]{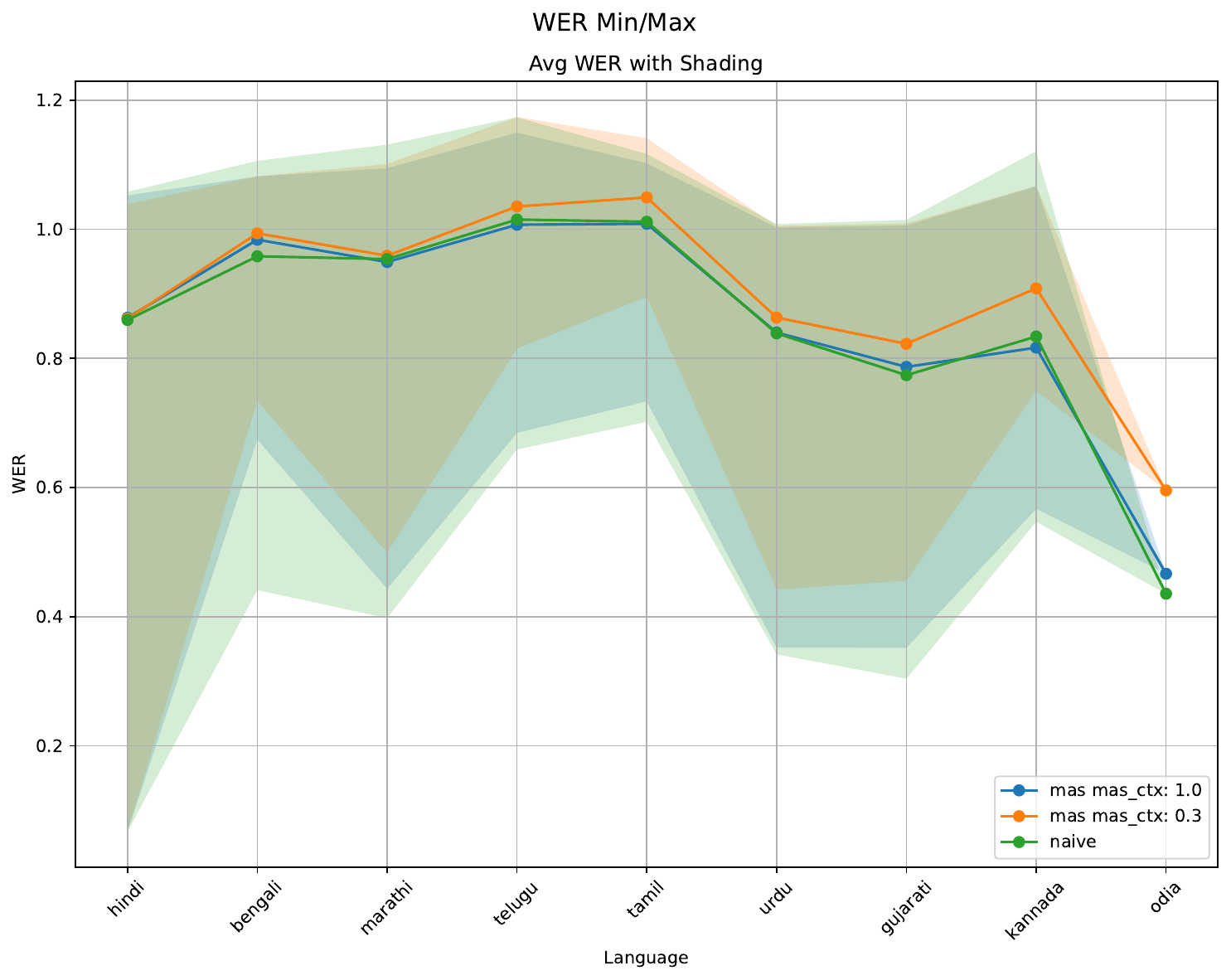}
\caption{MAS Shaded}
\end{subfigure}
\caption{MAS Ablation – Box and Shaded Plots}
\label{fig:mas_ablation_box_shaded}
\end{adjustwidth}
\end{figure*}


\begin{figure*}[ht]
\begin{adjustwidth}{-10cm}{-10cm}
\centering
\begin{subfigure}{0.4\textwidth}
\includegraphics[width=\linewidth]{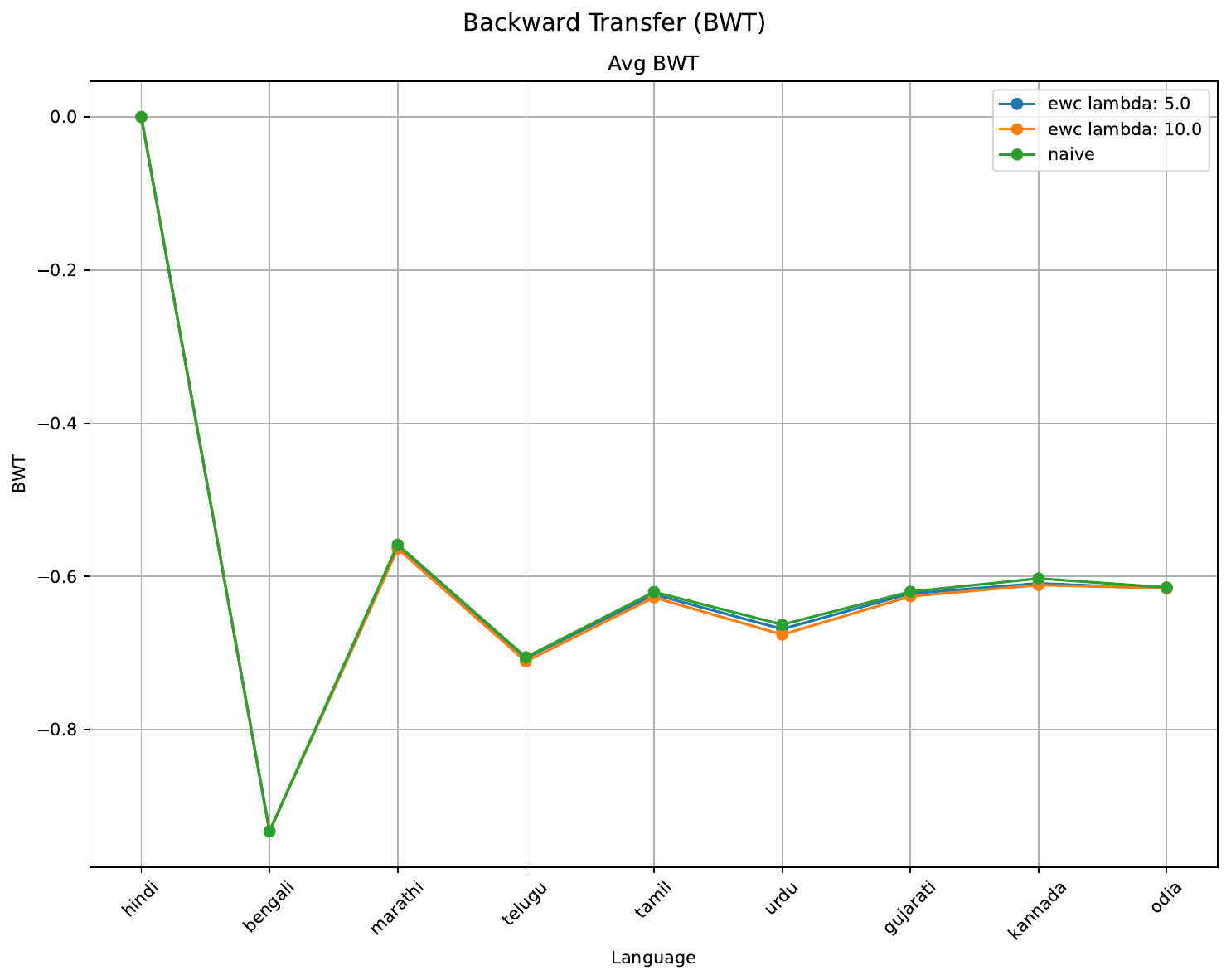}
\caption{EWC BWT}
\end{subfigure}
\begin{subfigure}{0.4\textwidth}
\includegraphics[width=\linewidth]{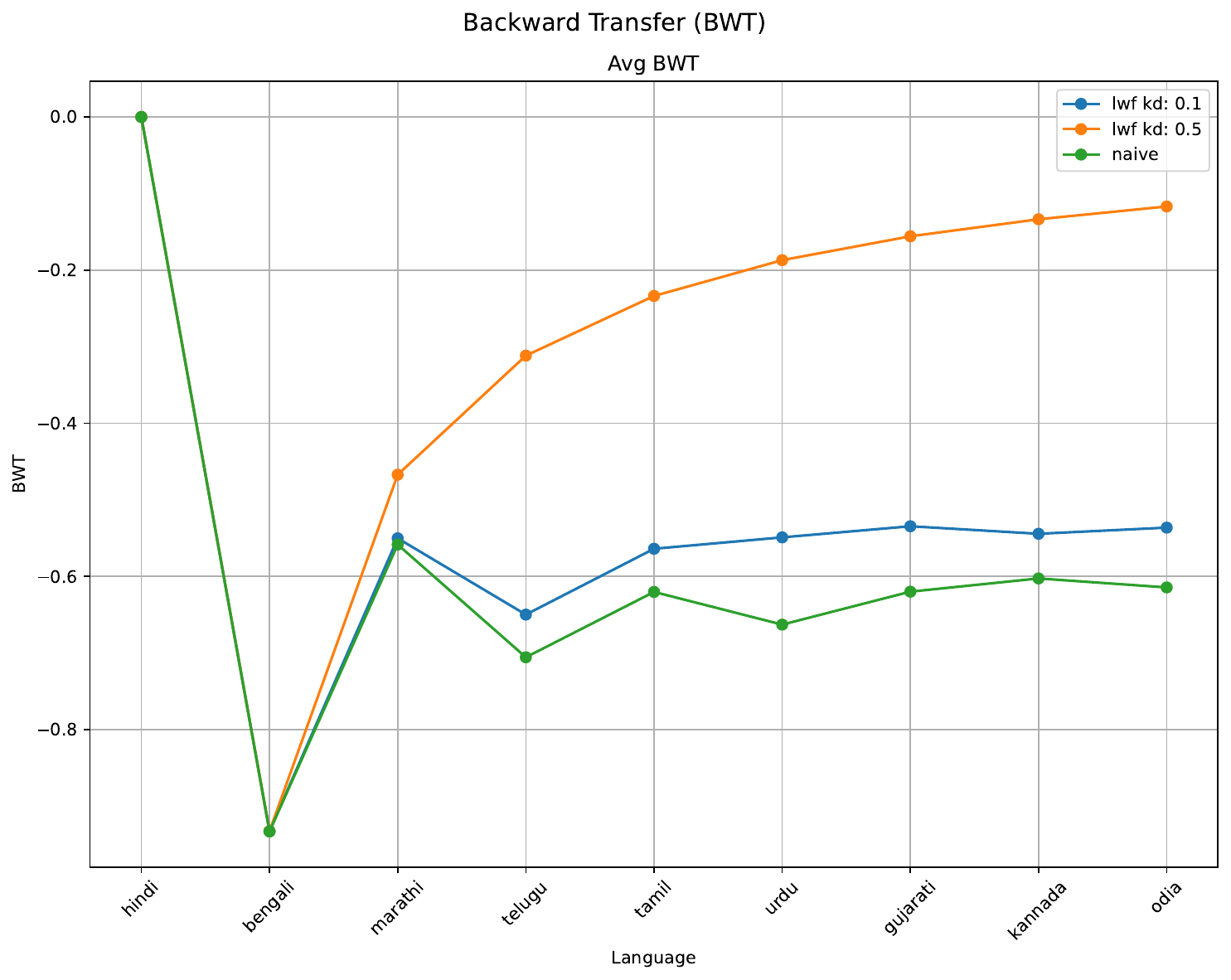}
\caption{LWF BWT}
\end{subfigure}
\begin{subfigure}{0.4\textwidth}
\includegraphics[width=\linewidth]{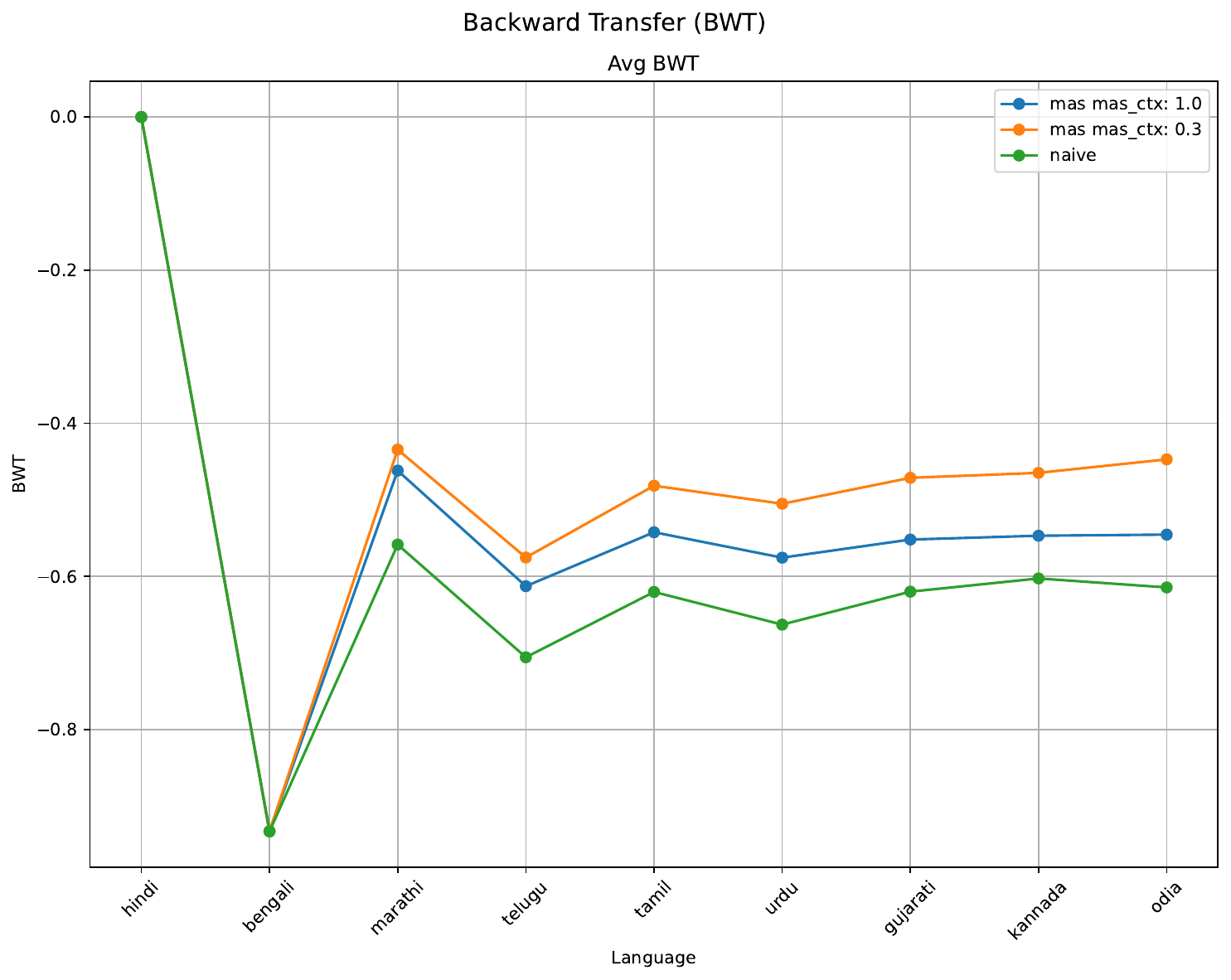}
\caption{MAS BWT}
\end{subfigure}
\caption{BWT Plots from EWC, LWF, and MAS Ablations}
\label{fig:bwt_all_ablation}
\end{adjustwidth}
\end{figure*}

\begin{figure*}[ht]
\begin{adjustwidth}{-10cm}{-10cm}
\centering
\begin{subfigure}{0.6\textwidth}
\includegraphics[width=\linewidth]{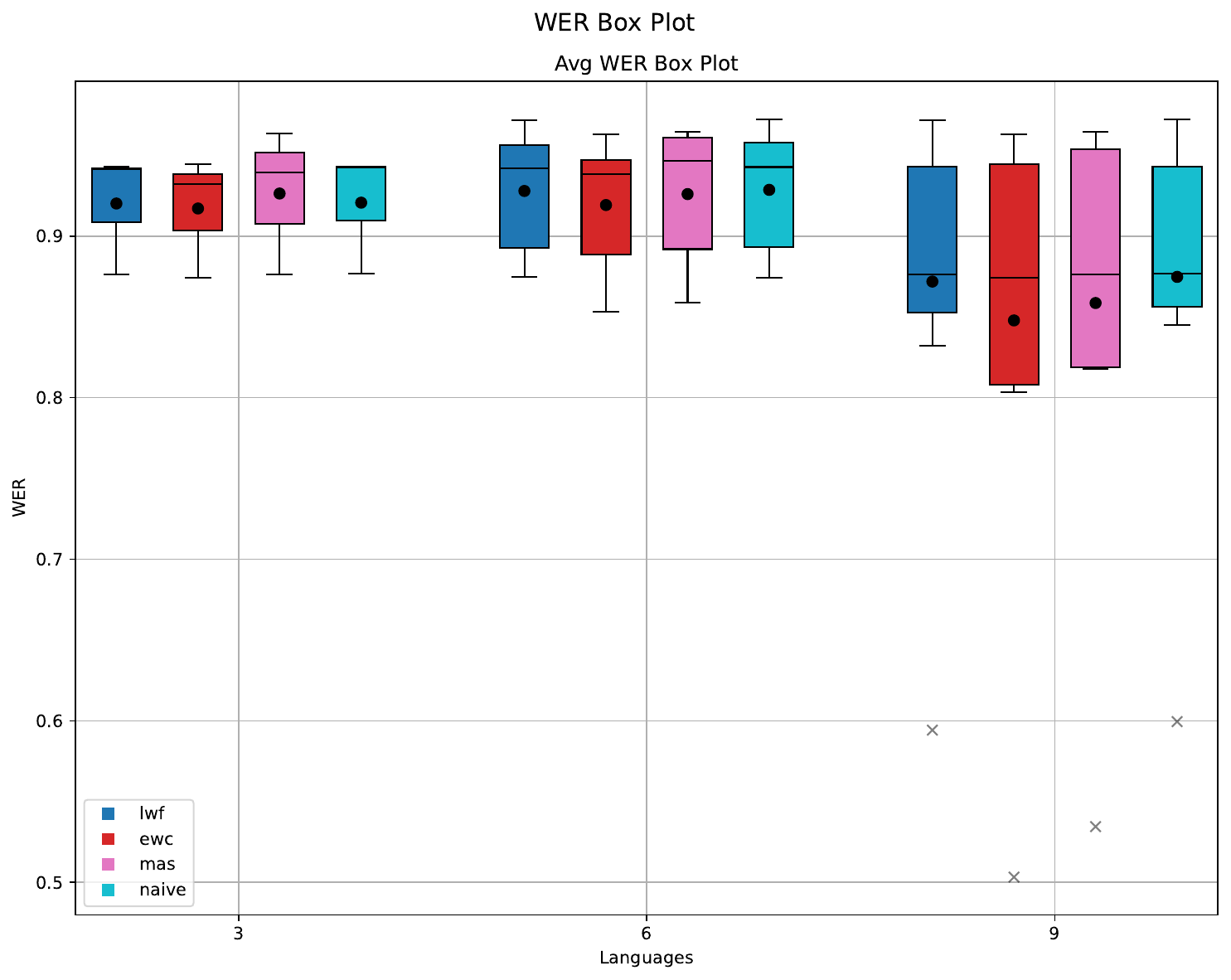}
\caption{RNN-T Box}
\end{subfigure}
\begin{subfigure}{0.6\textwidth}
\includegraphics[width=\linewidth]{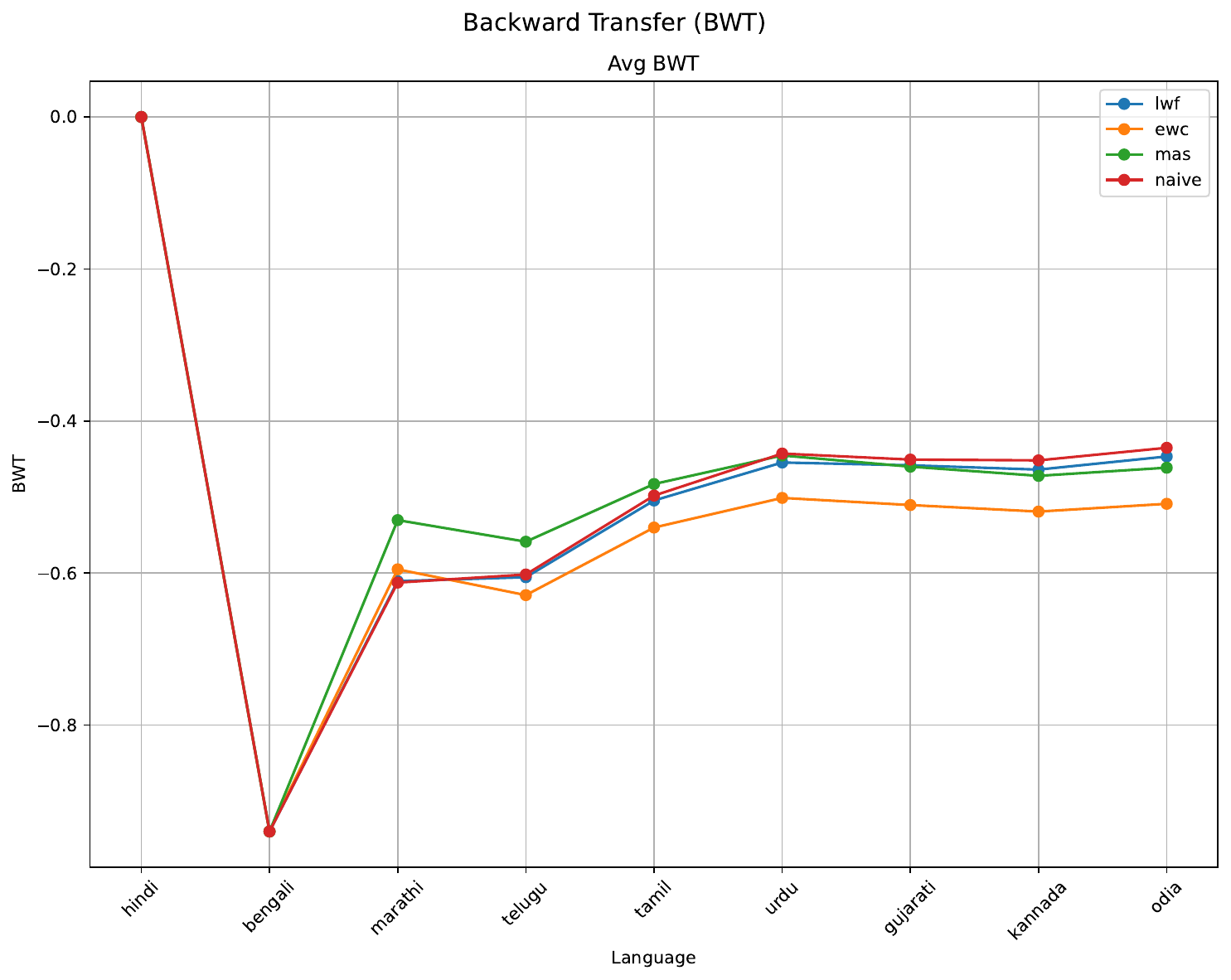}
\caption{RNN-T BWT}
\end{subfigure}
\caption{RNN-T Benchmark – Box and BWT Plots}
\label{fig:rnnt_benchmark_box_bwt}
\end{adjustwidth}
\end{figure*}

\begin{figure*}[ht]
\begin{adjustwidth}{-10cm}{-10cm}
\centering
\begin{subfigure}{0.6\textwidth}
\includegraphics[width=\linewidth]{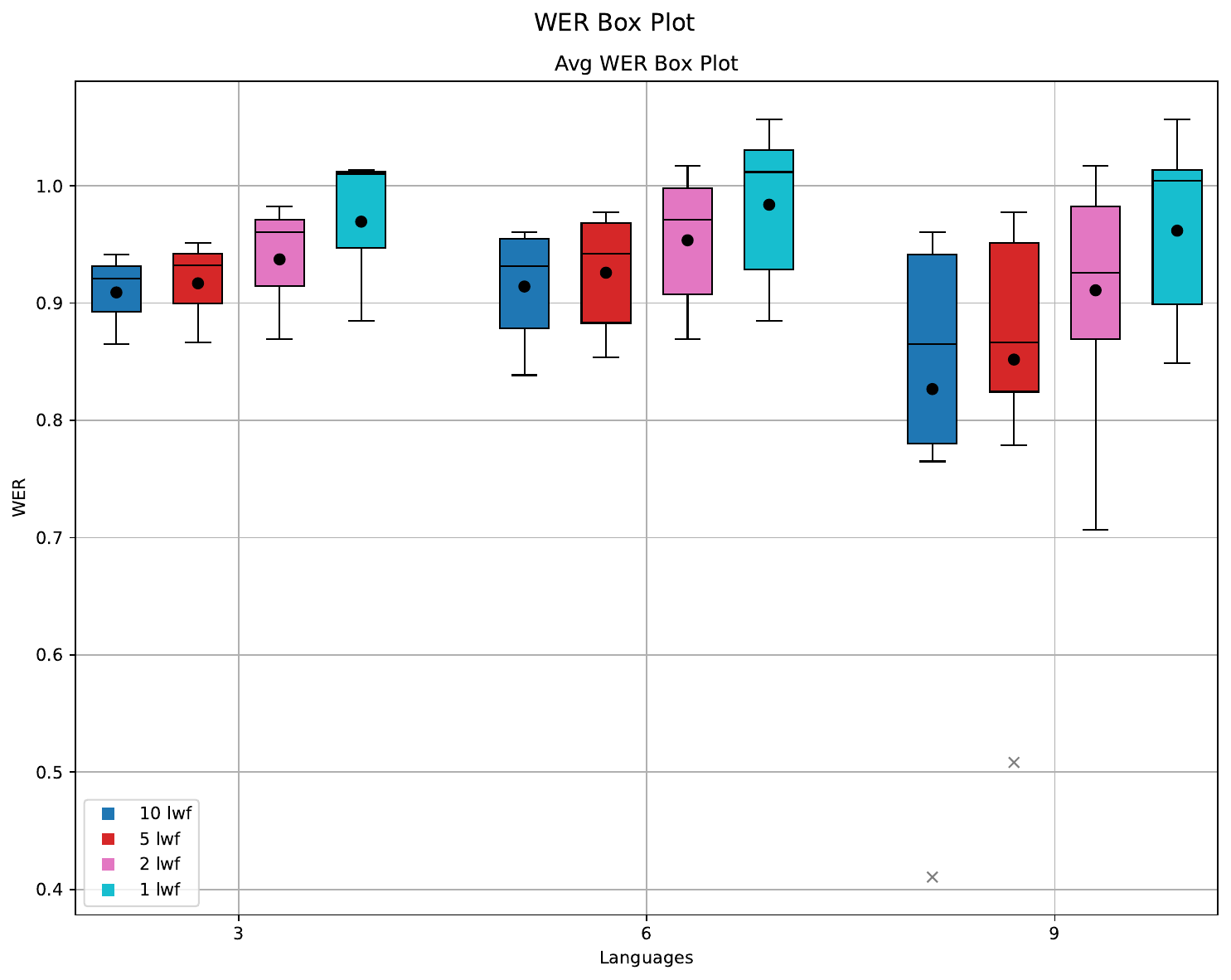}
\caption{LWF Epoch Box}
\end{subfigure}
\begin{subfigure}{0.6\textwidth}
\includegraphics[width=\linewidth]{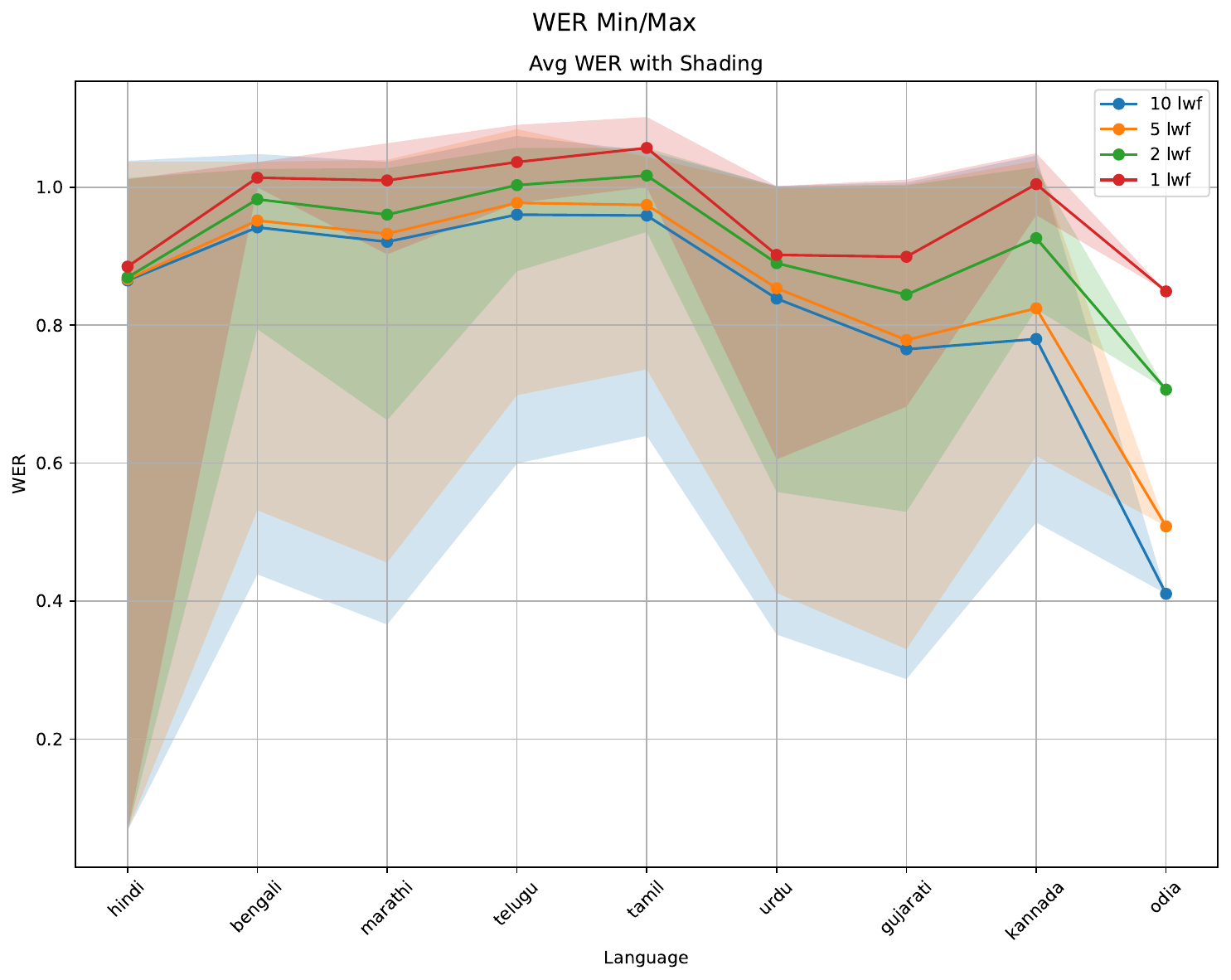}
\caption{LWF Epoch Shaded}
\end{subfigure}
\caption{LWF Epoch – Box and Shaded Plots}
\label{fig:lwf_epoch_box_shaded}
\end{adjustwidth}
\end{figure*}

\begin{figure*}[ht]
\begin{adjustwidth}{-10cm}{-10cm}
\centering
\begin{subfigure}{0.6\textwidth}
\includegraphics[width=\linewidth]{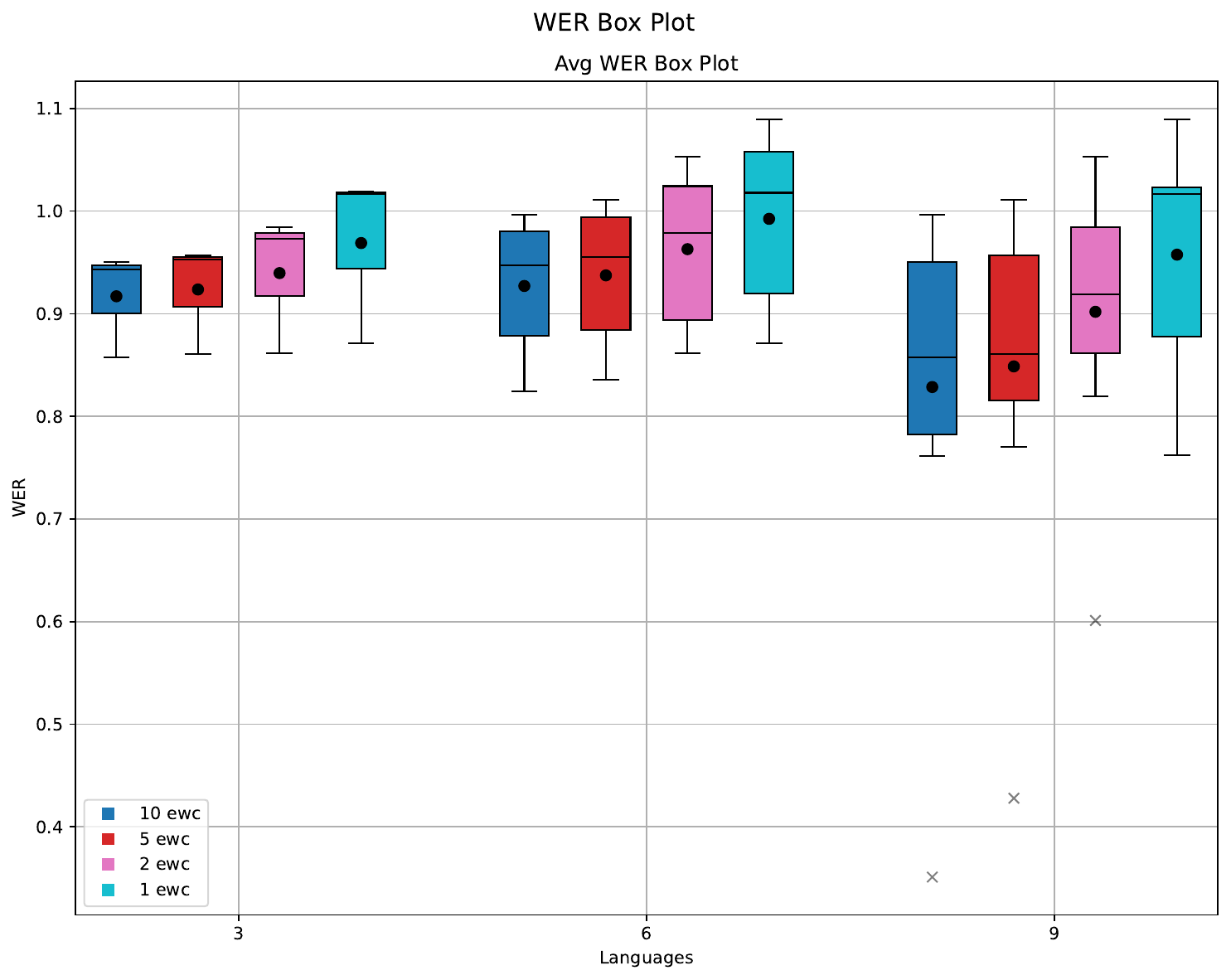}
\caption{EWC Epoch Box}
\end{subfigure}
\begin{subfigure}{0.6\textwidth}
\includegraphics[width=\linewidth]{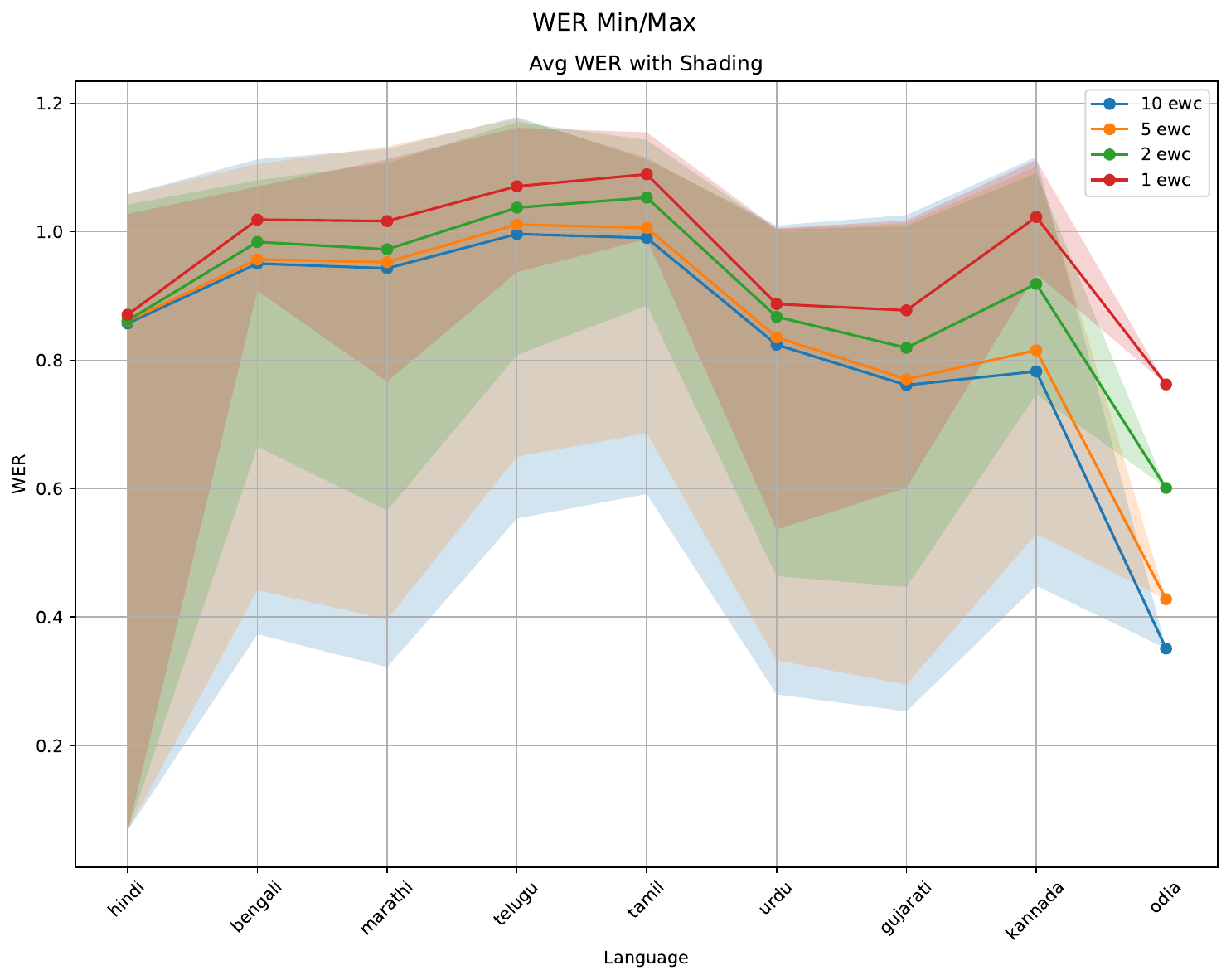}
\caption{EWC Epoch Shaded}
\end{subfigure}
\caption{EWC Epoch – Box and Shaded Plots}
\label{fig:ewc_epoch_box_shaded}
\end{adjustwidth}
\end{figure*}

\begin{figure*}[ht]
\begin{adjustwidth}{-10cm}{-10cm}
\centering
\begin{subfigure}{0.6\textwidth}
\includegraphics[width=\linewidth]{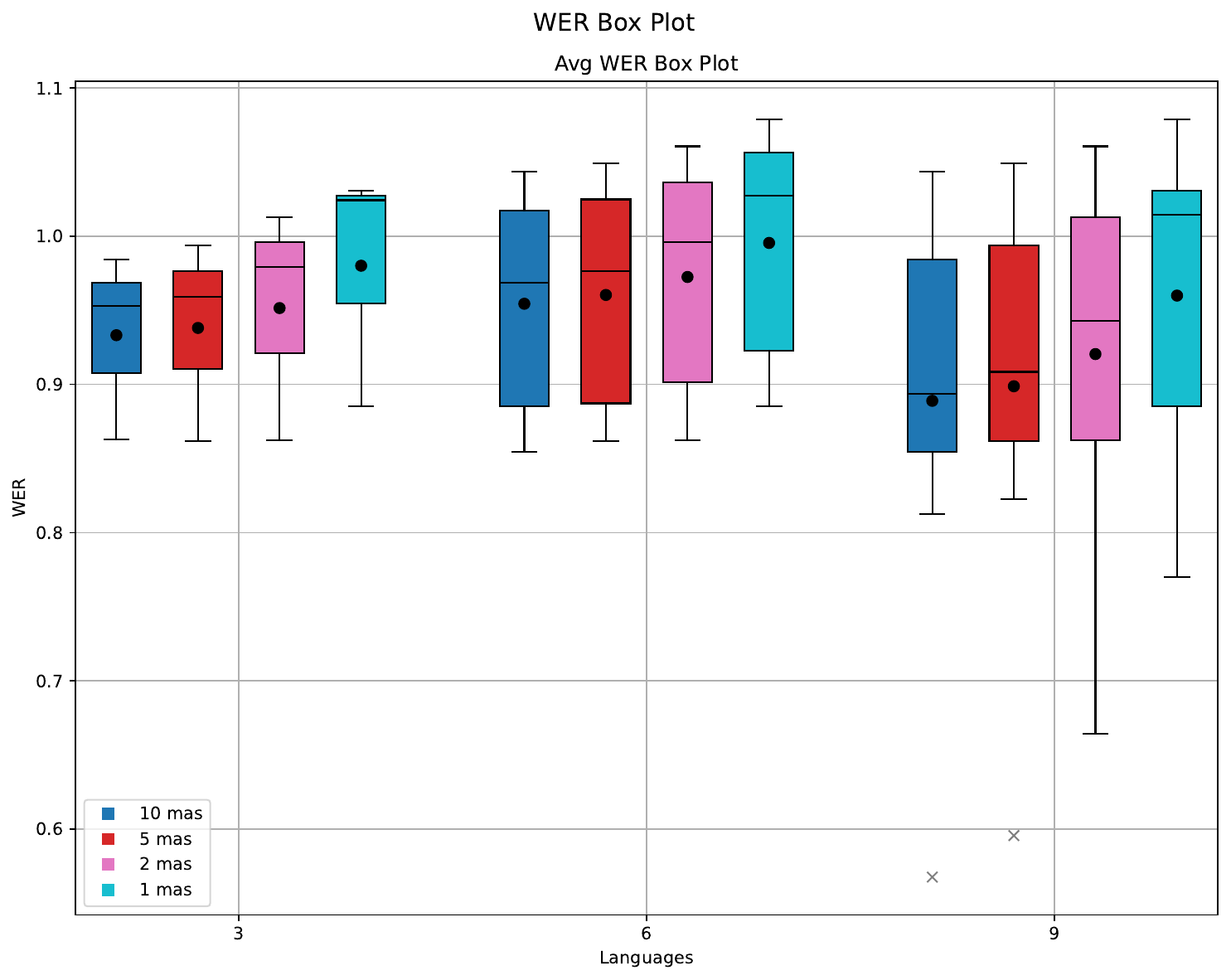}
\caption{MAS Epoch Box}
\end{subfigure}
\begin{subfigure}{0.6\textwidth}
\includegraphics[width=\linewidth]{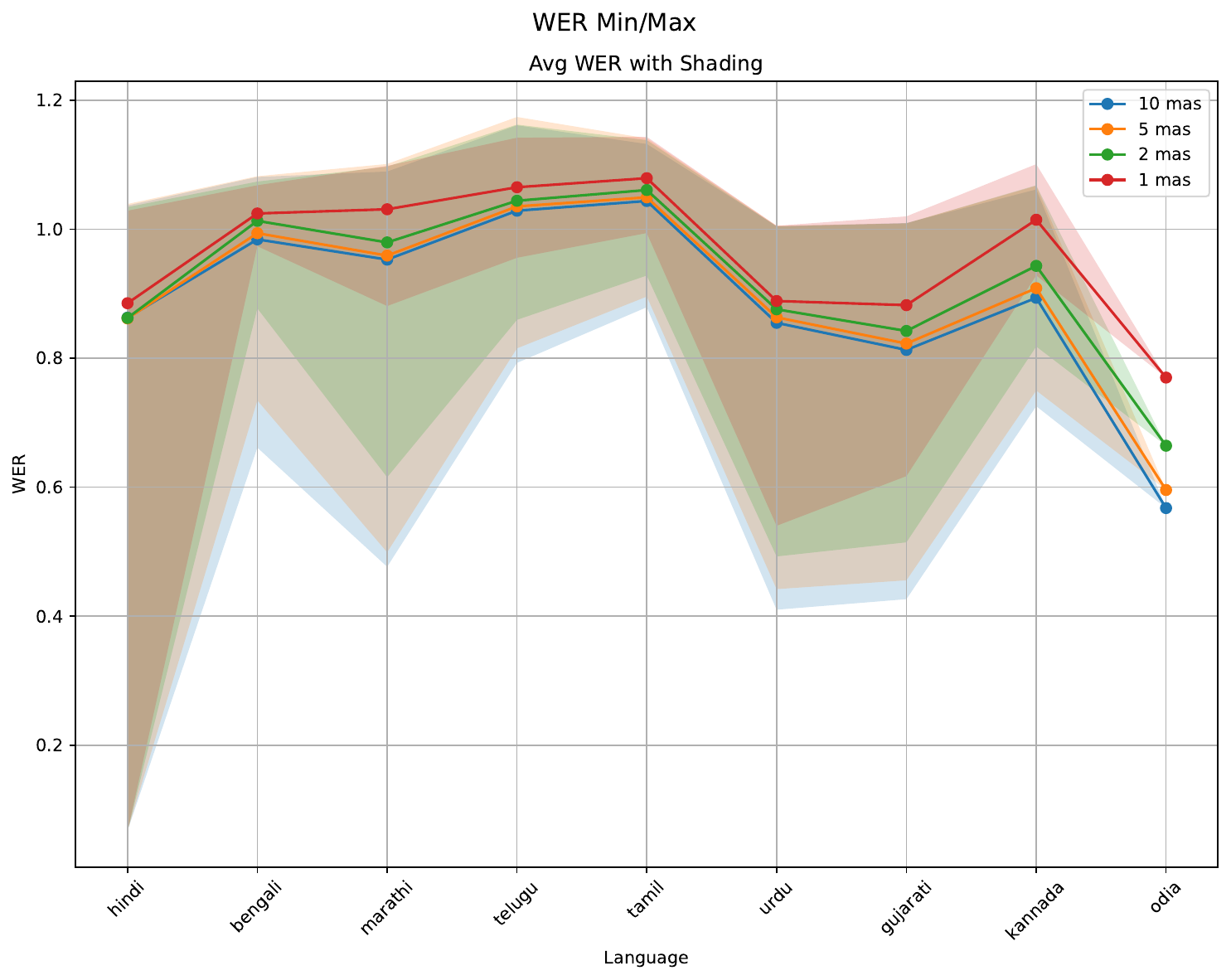}
\caption{MAS Epoch Shaded}
\end{subfigure}
\caption{MAS Epoch – Box and Shaded Plots}
\label{fig:mas_epoch_box_shaded}
\end{adjustwidth}
\end{figure*}

\begin{figure*}[ht]
\begin{adjustwidth}{-10cm}{-10cm}
\centering
\begin{subfigure}{0.6\textwidth}
\includegraphics[width=\linewidth]{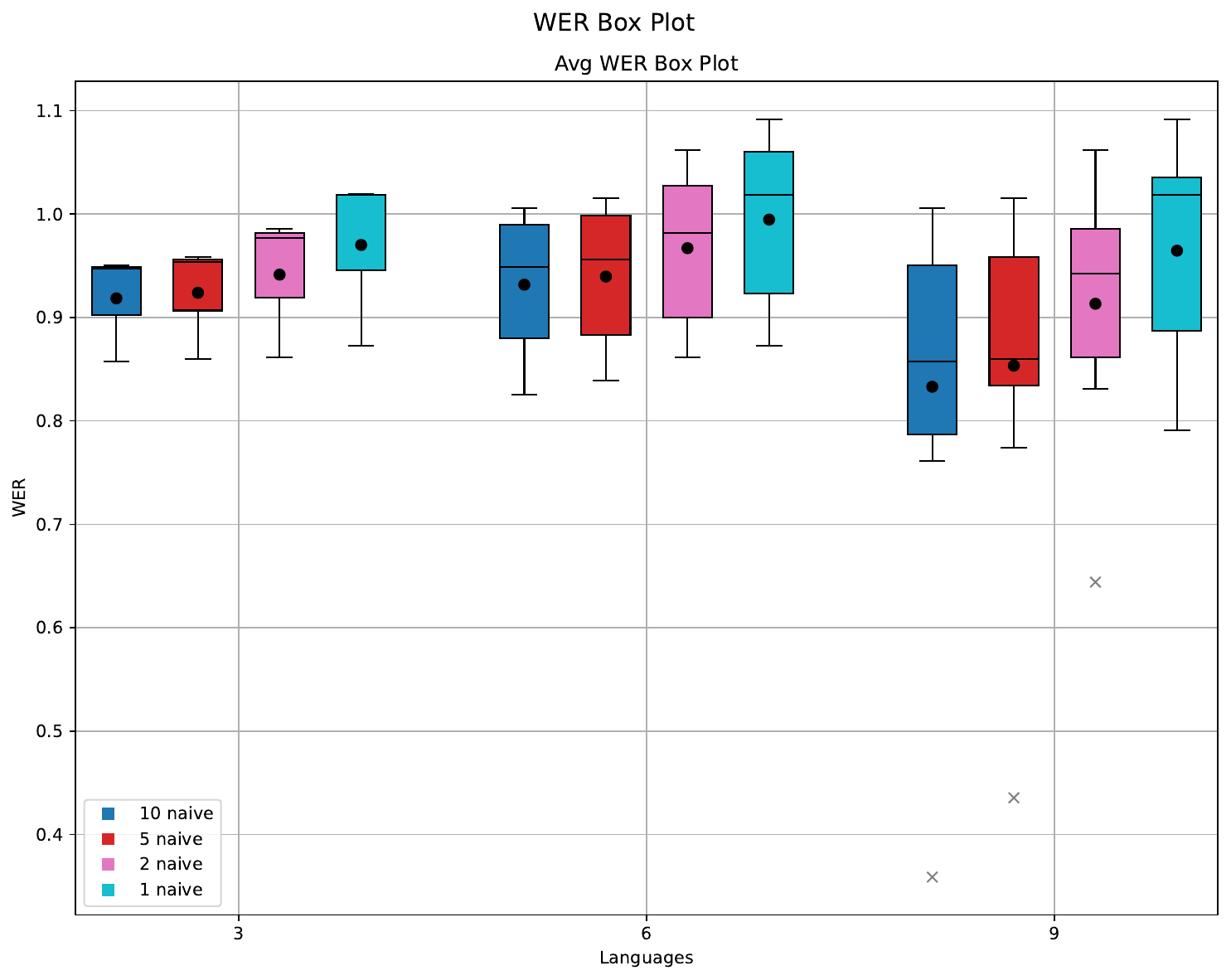}
\caption{Naive Epoch Box}
\end{subfigure}
\begin{subfigure}{0.6\textwidth}
\includegraphics[width=\linewidth]{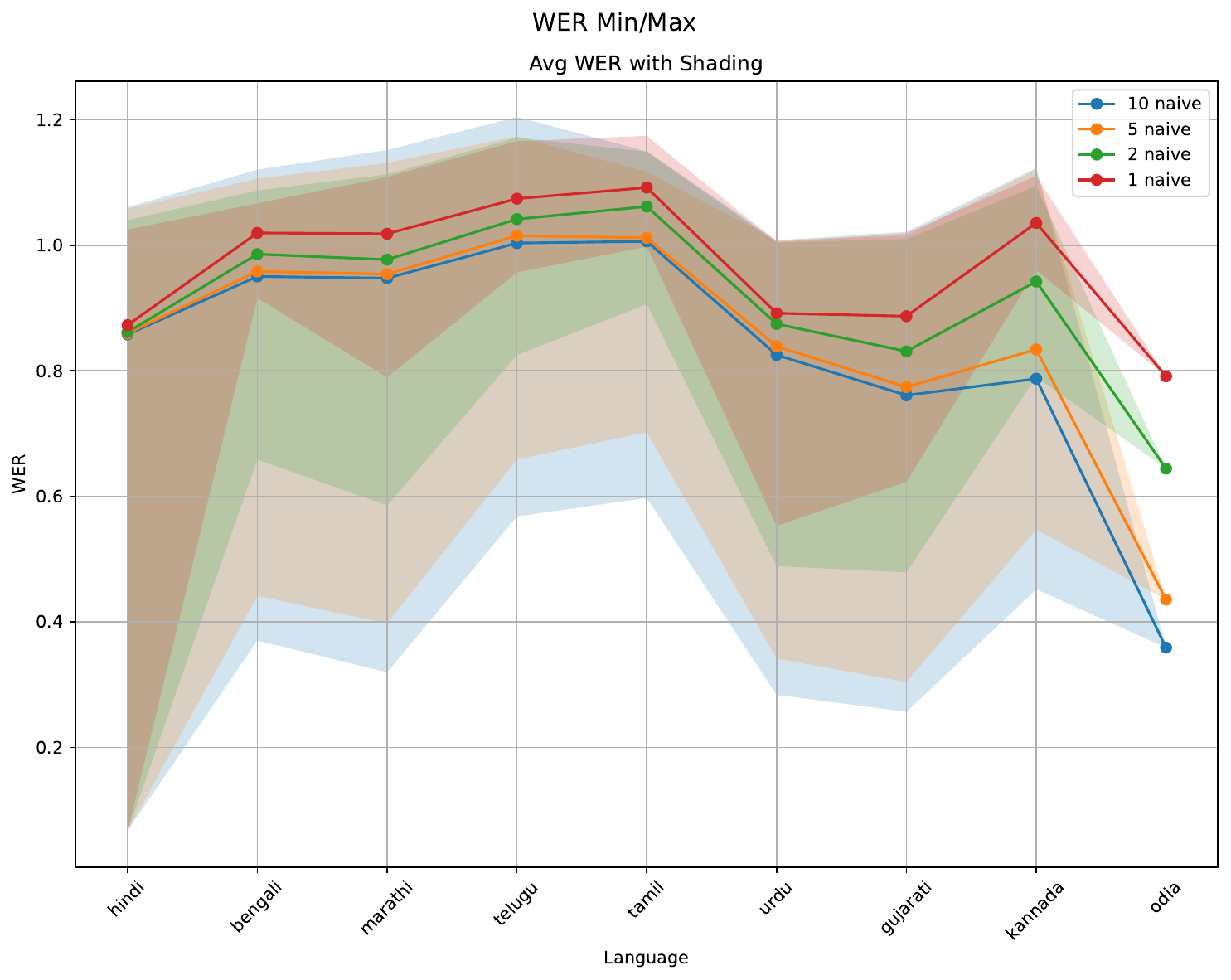}
\caption{Naive Epoch Shaded}
\end{subfigure}
\caption{Naive Epoch – Box and Shaded Plots}
\label{fig:naive_epoch_box_shaded}
\end{adjustwidth}
\end{figure*}

\begin{figure*}[ht]
\begin{adjustwidth}{-10cm}{-10cm}
\centering
\begin{subfigure}{0.48\textwidth}
\includegraphics[width=\linewidth]{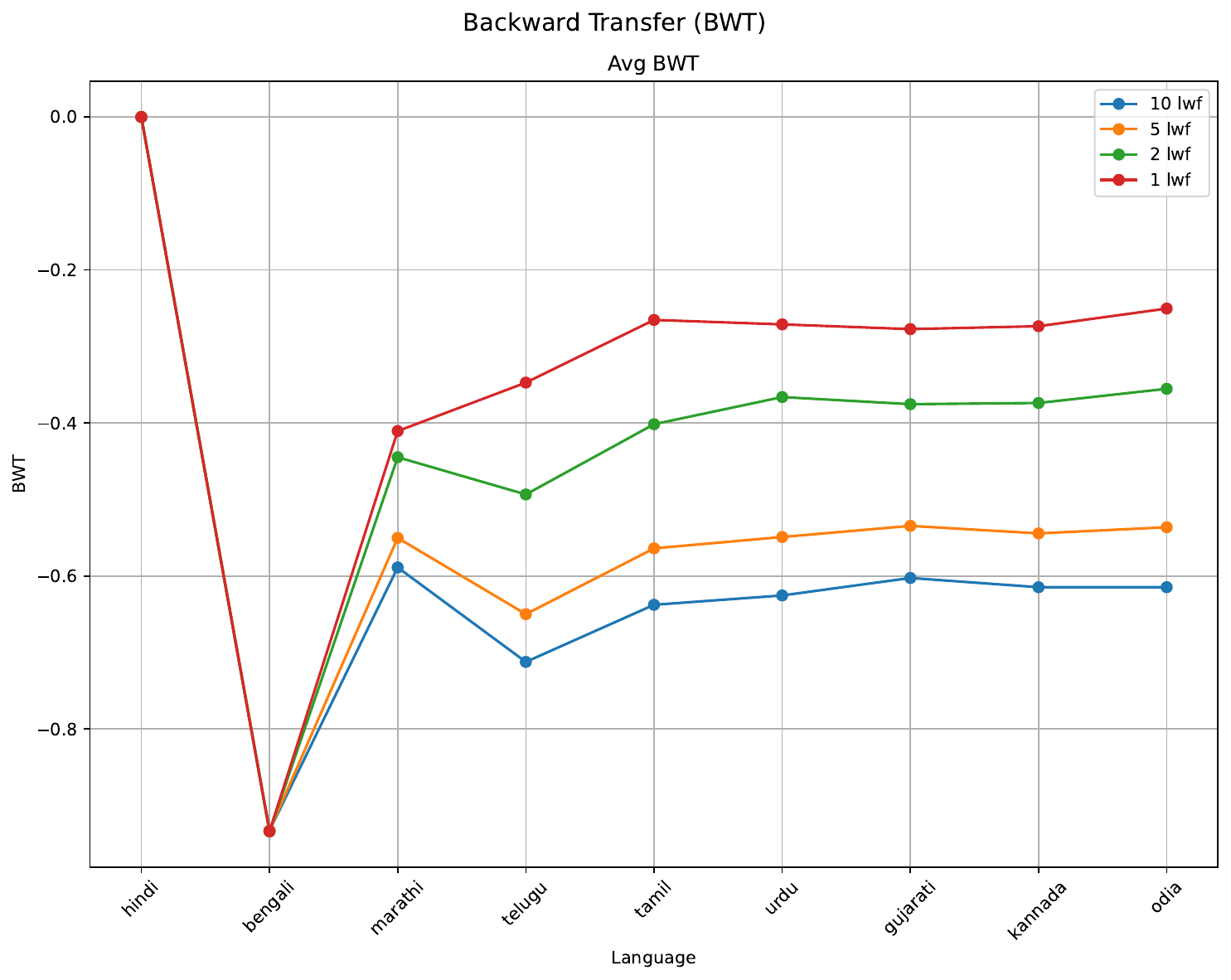}
\caption{LWF BWT}
\end{subfigure}
\begin{subfigure}{0.48\textwidth}
\includegraphics[width=\linewidth]{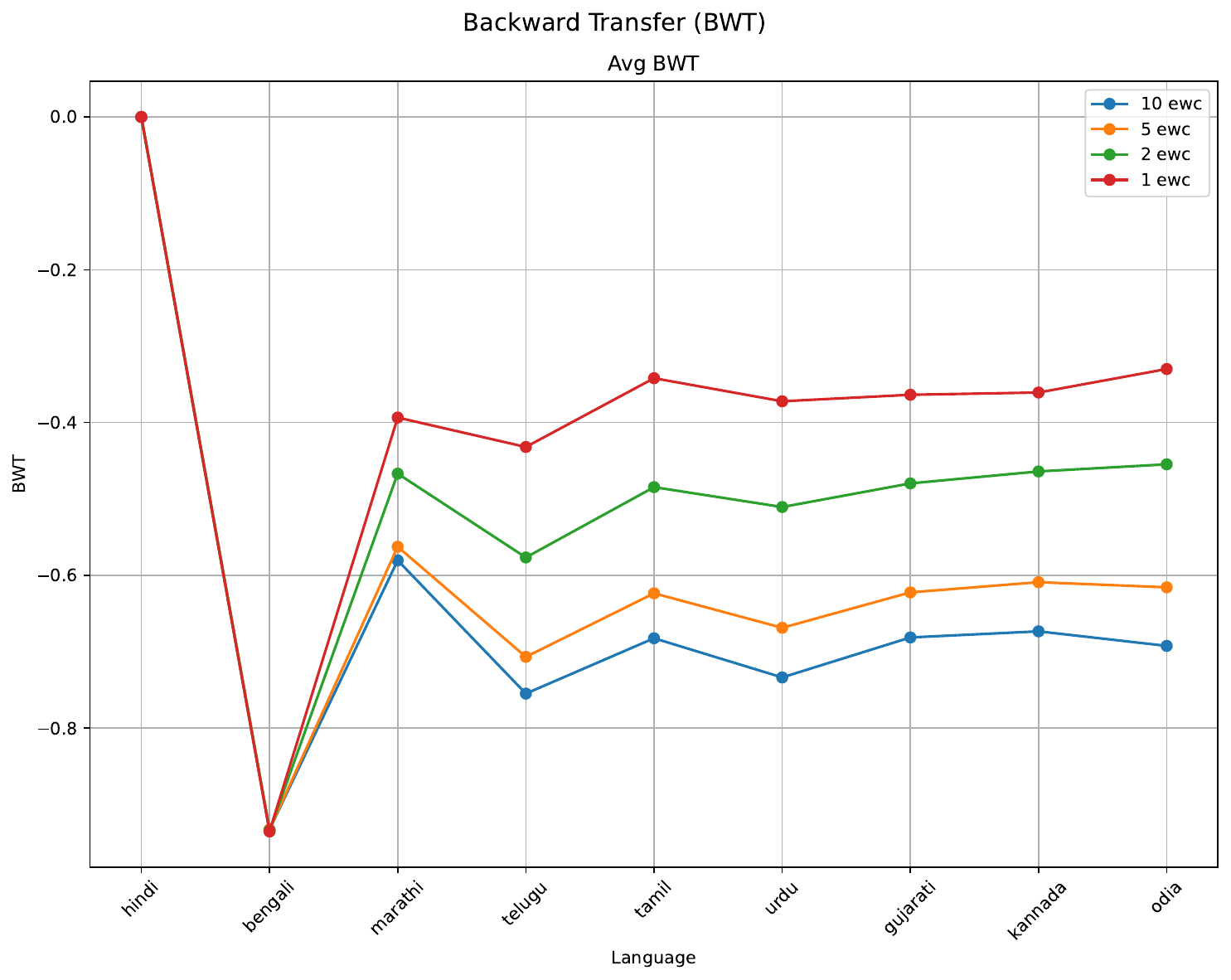}
\caption{EWC BWT}
\end{subfigure}

\vspace{0.5cm}

\begin{subfigure}{0.48\textwidth}
\includegraphics[width=\linewidth]{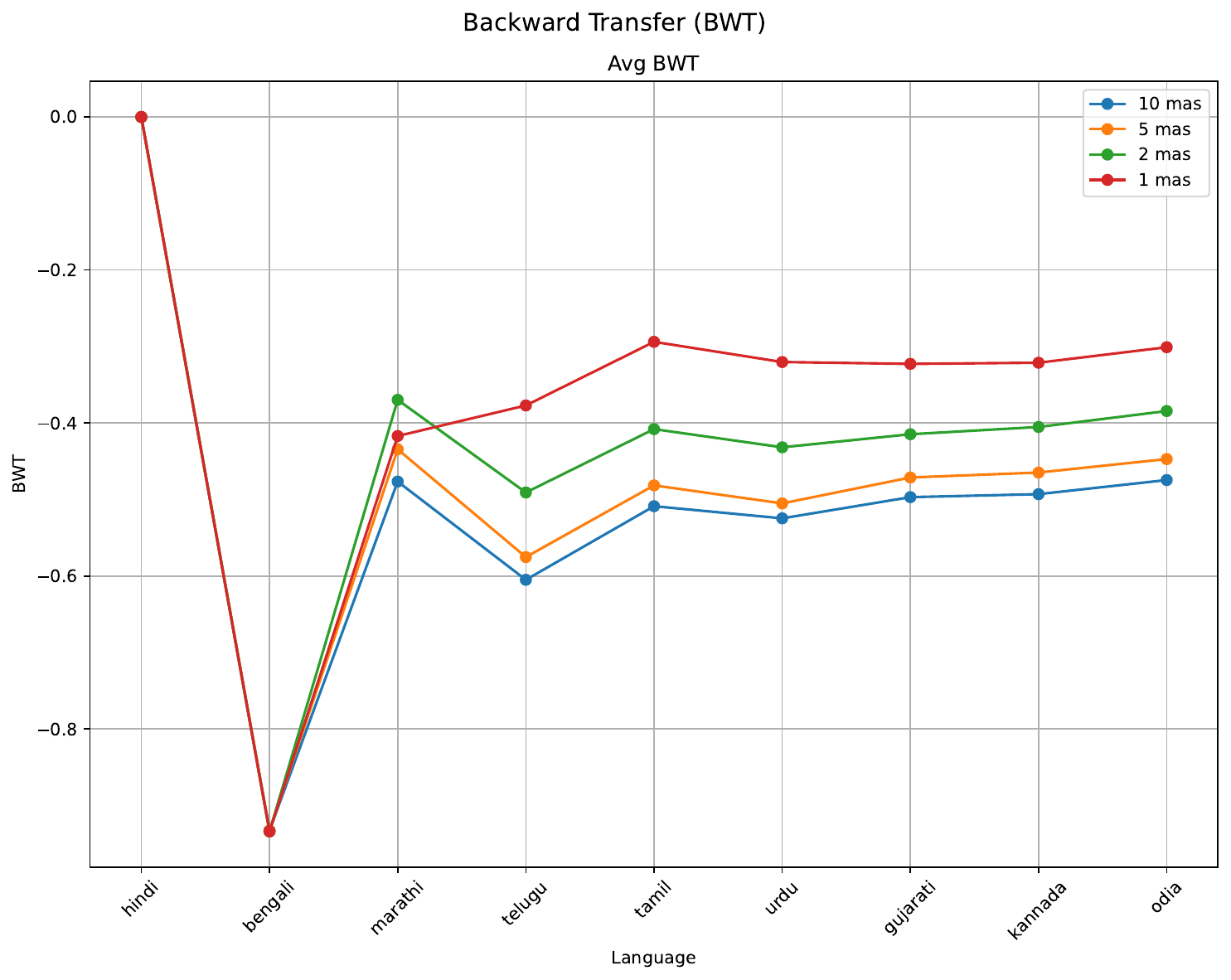}
\caption{MAS BWT}
\end{subfigure}
\begin{subfigure}{0.48\textwidth}
\includegraphics[width=\linewidth]{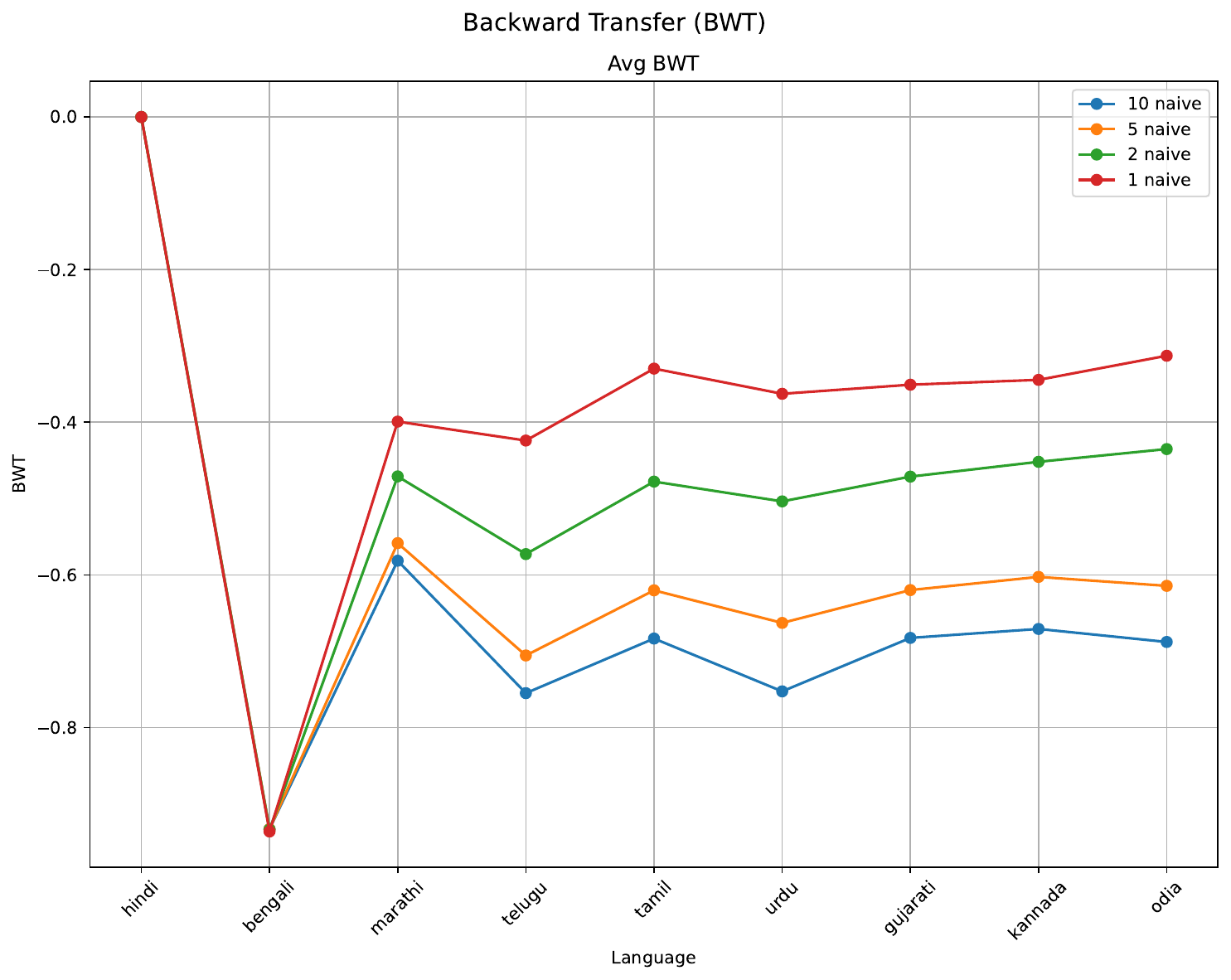}
\caption{Naive BWT}
\end{subfigure}

\caption{BWT Plots for Epoch-Wise Learning – LWF, EWC, MAS, and Naive}
\label{fig:bwt_epoch_grid}
\end{adjustwidth}
\end{figure*}



\end{document}